\documentclass[accepted]{uai2026} 
                        
\usepackage[american]{babel}

\usepackage{natbib} 
\usepackage{mathtools} 
\usepackage{booktabs} 
\usepackage{tikz} 

\usepackage{amsmath,amsfonts,bm}

\def\1{\bm{1}}

\def\rvw{{\mathbf{w}}}
\def\rvx{{\mathbf{x}}}
\def\rvy{{\mathbf{y}}}
\def\rvz{{\mathbf{z}}}

\def\vf{{\bm{f}}}

\def\vh{{\bm{h}}}

\def\vm{{\bm{m}}}

\def\vp{{\bm{p}}}

\def\vv{{\bm{v}}}

\def\vx{{\bm{x}}}
\def\vy{{\bm{y}}}
\def\vz{{\bm{z}}}

\def\mA{{\mathbf{A}}}

\def\mC{{\mathbf{C}}}

\def\mH{{\mathbf{H}}}
\def\mI{{\mathbf{I}}}

\def\mK{{\mathbf{K}}}

\def\mM{{\mathbf{M}}}

\def\mT{{\mathbf{T}}}
\def\mU{{\mathbf{U}}}
\def\mV{{\mathbf{V}}}

\newcommand{\E}{\mathbb{E}}

\newcommand{\R}{\mathbb{R}}

\newtheorem{definition}{Definition}

\usepackage{amsmath}
\usepackage{amssymb}
\usepackage{bm}
\usepackage{multirow}
\usepackage{graphicx}
\usepackage{tabularx}
\usepackage{algorithm}
\usepackage{algorithmic}
\usepackage{url}

\title{Deep Spectral Learning of Embedded Latent Transfer Operators\\for Stochastic Dynamical Systems}

\author[1,2]{\href{mailto:<ryogo.tanaka@ist.osaka-u.ac.jp>}{Ryogo Tanaka}{}}
\author[1,2]{\href{mailto:<kawahara@ist.osaka-u.ac.jp>}{Yoshinobu Kawahara}{}}

\affil[1]{%
    Graduate School of Information Science and Technology, The University of Osaka \\
    Osaka \\
    Japan
}
\affil[2]{%
    Center for Advanced Intelligence Project, RIKEN\\
    Tokyo\\
    Japan
}

\begin{document}


\newtheorem{assumption}[theorem]{Assumption}
\newtheorem{example}[theorem]{Example}

\maketitle

\begin{abstract}
We propose a spectral learning method for stochastic nonlinear dynamical systems represented with embedded latent transfer operators in deep feature spaces. We instantiate the method as Deep Spectral Encoder (DSE), an operator-based latent state-space model in which a time-invariant neural encoder implements learnable nonlinear feature maps from observations, and these features define Markovian latent states whose temporal evolution and observation mapping are described by the transfer and observation operators, respectively. Functional canonical correlation analysis in a learnable Galerkin-projected feature space provides state coordinates from past and future observations, and the two linear operators are estimated on the state coordinates as ridge-regularized closed-form solutions that coincide with Galerkin projections of the associated covariance operators. On this representation, we generalize sequential Bayesian filtering 
and Koopman spectral mode decomposition in feature space. Experiments on several scenarios show stable and superior performance with sequential Bayesian filtering and dynamic mode decomposition baselines even under noise and partial observability.

\end{abstract}

\section{Introduction}
\label{sec:intro}

Understanding and forecasting the evolution of stochastic nonlinear dynamics using time-series data is a central yet technically demanding problem across science and engineering. Real-world time series data often appear non-stationary in the observation space due to nonlinear evolution, observational noise, and partial observability, so that simple linear models fail to capture the underlying structure.

A classical approach to forecasting in dynamical systems is sequential Bayesian filtering, typified by the Kalman filter \citep{Kal60} and its nonlinear extensions, such as the extended and unscented Kalman filters, and kernel Kalman filters \citep{Anderson1979, doucetfg01, GKN19}. When state space models are learned from data, common strategies infer latent state trajectories using expectation maximization or variational approximations, or learn deep latent state models with amortized inference \citep{krishnan2015, BPG+19}. Continuous-time latent stochastic dynamics have also been modeled through parametric or variational latent SDE formulations \citep{hasan2022, duncker19}. Another line of work adopts {\em spectral learning} to identify a latent Markov representation directly from empirical moments of past and future observations, avoiding iterative latent state inference. In statistics and control, stochastic realization and subspace identification construct predictive state coordinates via canonical correlation analysis (CCA) between past and future blocks and estimate dynamics by regression \citep{Aka75, Desai1985, katayama2005subspace}. In machine learning, spectral algorithms were developed for hidden Markov models \citep{Hsu09} and extended to nonlinear settings through kernel-based stochastic realization and Hilbert space embeddings \citep{Kawahara06, Song2010,nry2025elto}. These estimators are closed-form and statistically well behaved, but their effectiveness depends on feature representations that are usually fixed in advance.

We propose Deep Spectral Learning (DSL), a method for stochastic dynamical systems that learns embedded latent transfer operators in finite-dimensional deep feature spaces while retaining an operator-based state space formulation with closed-form operator estimation. As a concrete instantiation of DSL, we introduce Deep Spectral Encoder (DSE), in which a time-invariant neural encoder maps observation windows to block-wise features and functional CCA in a learnable Galerkin-projected feature space yields predictive state coordinates. On these coordinates, the transfer and observation operators are estimated as ridge-regularized closed-form solutions via a deep-feature two-stage procedure \citep{Xu2021}, and these solutions coincide with Galerkin projections of associated conditional covariance operators onto the subspaces spanned by the learned features. This representation generalizes sequential Bayesian filtering in feature space by propagating and updating embedded state laws with the learned operators, and it also generalizes Koopman spectral analysis via eigen-decomposition of the learned transfer operator, respectively in closed-forms. Experiments on several scenarios demonstrate robust forecasting and spectrum recovery under noise and partial observability, with superior performance to the baseline methods.

In this framework, Hilbert-space embeddings and conditional covariance operators serve as the standard population-level operator view used to interpret the estimators, while the role of DSL is to make the finite-dimensional representation spaces used by spectral realization learnable. DSE specifies these spaces by neural encoder and dictionary maps: the encoder and block-feature maps define the past--future CCA problem that yields predictive state coordinates, while the state and observation dictionaries define the spaces spanned by these functions for closed-form transfer and observation operator estimation. It differs from the kernel-based realization in ELTO~\citep{nry2025elto}, whose empirical representation is expressed through kernel evaluations over samples in an RKHS; DSE instead learns explicit finite-dimensional Galerkin coordinates and performs operator regression in these realized coordinates. The feature-space filtering and Koopman-style spectral analysis developed below are downstream uses of this learned operator representation.


The remainder of this paper is organized as follows. Section~\ref{sec:background} reviews the mean embeddings, covariance operators, and deep-feature-based two-stage estimation of linear operators. Section~\ref{sec:Op-bsd-SSM} develops the operator-based state-space formulation and stochastic realization with deep features, and Section~\ref{sec:prop} presents the DSE model and its learning algorithm. We develop sequential state estimation and spectral mode decomposition with our operator-based representation, respectively, in Section~\ref{sec:application:seq} and in Section~\ref{sec:application:mode}, with experimental results, 
and Section~\ref{sec:cclsn} concludes the paper. The code is available at \url{https://github.com/uosaka-mlsyslab/DeepSpectralEncoder}.

\section{Background}
\label{sec:background}

\subsection{Hilbert Space Embedding of Conditional Distributions}
\label{subsec:embed-CD}


Embedding probability laws, including conditional ones, into Hilbert spaces allows us to realize expectations and conditional expectations as inner products and linear operators on feature spaces. This subsection recalls the corresponding Hilbert-space embedding framework for random variables, including mean elements that represent marginal laws and the covariance, cross-covariance, and conditional covariance operators \citep{baker1973,Hsing2015}. These constructions are used as standard background and provide the population-level operator view for the finite-dimensional estimators introduced later.


Let $\rvx$ and $\rvy$ be random variables taking values $\vx\in\mathbb{X} $ and $ \vy\in\mathbb{Y} $, where $\mathbb{X}$, $\mathbb{Y}$ are real vector spaces, with corresponding marginal distributions  $P_\rvx$, $P_\rvy$ and joint distribution $P_{\rvx\rvy}$.
Let $\mathbb{H}$ and $\mathbb{G}$ be separable Hilbert spaces, and consider feature maps $\phi_x:\mathbb{X}\rightarrow \mathbb{H}$, $\psi_y:\mathbb{Y}\rightarrow \mathbb{G}$ such that $\E\left[\|\phi_x(\rvx)\|^2\right]<\infty$ and $\E\left[\|\psi_y(\rvy)\|^2\right]<\infty$. 
We represent the marginal distributions of $\rvx$ and $\rvy$ in feature spaces by their mean elements. The mean element of $\rvx$ is defined by
\begin{align*}
\mu_\rvx := \mathbb{E}\big[\phi_x(\rvx)\big] = \int \phi_x(\vx)\, dP_\rvx(\vx) \in \mathbb H,
\end{align*}
and 
$\mu_\rvy \in \mathbb G$ is defined analogously 
\citep{SGSS07}. The square-integrability assumptions ensure that these expectations are well-defined as Hilbert space elements. This construction is called a mean embedding and it satisfies $\mathbb{E}[f(\rvx)]=\mathbb{E}[\langle f,\phi_x(\rvx)\rangle_{\mathbb H}] = \langle f,\mu_\rvx\rangle_{\mathbb H}$ for all $f \in \mathbb H$. Under suitable conditions on the feature map, the mean embedding is injective and therefore characterizes the underlying distribution \citep{Sriperumbudur2008}.

Mean embeddings provide a first-order summary of the marginal distributions in the feature spaces. To describe second-order relationships in the feature space, we introduce covariance and cross-covariance operators, defined as expectations of rank-one operators induced by the feature maps \citep{baker1973}.
\begin{definition}
    Recall that $\phi_x(\rvx)\in\mathbb{H}$ and $\psi_y(\rvy)\in\mathbb{G}$ denote the feature maps of $\rvx$ and $\rvy$ introduced above, and assume they are square-integrable. Then, the covariance operator $\mathcal{C}_\rvx:\mathbb{H} \rightarrow \mathbb{H}$ and cross-covariance operator $\mathcal{C}_{\rvy\rvx}:\mathbb{H} \rightarrow \mathbb{G}$ are defined by
    \begin{align*}
        \mathcal{C}_\rvx &:=\int \phi_x(\bm{x}) \otimes \phi_x(\bm{x}) dP_\rvx(\bm{x}), \\
        \mathcal{C}_{\rvy\rvx} &:=\int \psi_y(\bm{y}) \otimes \phi_x(\bm{x}) dP_{\rvx\rvy}(\bm{x}, \bm{y}), 
    \end{align*}
\end{definition}
where the rank-one operator $a\otimes b :\mathbb{H}\rightarrow \mathbb{G}$ is defined by $(a \otimes b ) w := \langle w, b\rangle_\mathbb{H}\, a$ for all $b,\,w\in \mathbb{H}$ and $a\in\mathbb{G}$. For any $f \in \mathbb{H}$ and $g\in \mathbb{G}$,
\begin{align*}
    \E_{\rvx\rvy}[f(\rvx)g(\rvy)]  = \left<\mathcal{C}_{\rvy\rvx}\,f, g\right>_{\mathbb{G}} =\left<f, \mathcal{C}_{\rvy\rvx}^* \,g\right>_{\mathbb{H}},
\end{align*}
which is the standard representation of cross-covariance operators for Hilbert-space-valued random variables.

In the integrable function space, conditional expectation can be viewed as the orthogonal projection onto the closed subspace generated by $\phi_x(\rvx)$ \citep{Hsing2015}. Restricting to linear predictors of the form $\mathcal{A}\,\phi_x(\rvx)$ with $\mathcal{A}\in\mathcal{L}(\mathbb{H}, \mathbb{G})$, where $\mathcal{L}(\mathbb{H}, \mathbb{G})$  denotes the space of bounded linear operators from $\mathbb{H}$ to $\mathbb{G}$, leads to the regularized least-squares problem with a regularization parameter $\lambda>0$
\begin{align*}
    \mathcal{A}^* := \underset{\mathcal{A}\in\mathcal{L}(\mathbb{H}, \mathbb{G})}{\mathrm{argmin}} \E\left[ \|\psi_y(\rvy)-\mathcal{A}\phi_x(\rvx)\|^2_\mathbb{G}\right] + \lambda \|\mathcal{A}\|_{\mathrm{HS}}^2,
\end{align*}
where $\|\cdot\|_{\mathrm{HS}}$ is the Hilbert-Schmidt norm.  The corresponding normal equations yield the unique solution as $\mathcal{A}^* = \mathcal{C}_{\rvy\rvx} (\mathcal{C}_{\rvx} + \lambda\, \mathcal{I})^{-1}$,
where $\mathcal{I}$ is the identity operator. The operator $\mathcal{A}^{*}$ coincides with the population optimal linear predictor of $\psi_y(\rvy)$ from $\phi_x(\rvx)$ in the mean-squared sense \citep{grunewalder2012}. Motivated by this characterization and by the operator-theoretic treatment of conditional covariance based on $\mathcal{C}_{\rvx},\,\mathcal{C}_{\rvy\rvx}$ in \citep{Fukumizu2004,Song2009}, this leads to the following definition:
\vspace{-5pt}
\begin{definition}
    We define $\mathcal{C}_{\rvy\mid\rvx} :=  \mathcal{C}_{\rvy\rvx} (\mathcal{C}_{\rvx} + \lambda\, \mathcal{I})^{-1}$ and refer to $\mathcal{C}_{\rvy\mid\rvx}:\mathbb H\to \mathbb G$ as the (regularized) conditional covariance operator.
\end{definition}
\vspace{-5pt}
This operator is the ridge-regularized population-optimal linear predictor of $ \psi_y(\rvy)$ from $\phi_x(\rvx)$, and thus serves as a linear representation of conditional expectations in feature space. In practice, such conditional covariance operators are approximated on some 
finite-dimensional subspaces. 

\subsection{Two-Stage Estimation with Deep Features}


In the previous section, we introduced operators such as covariance and conditional covariance operators acting on feature spaces. To obtain learnable finite-dimensional representations of these operators, we choose neural feature maps that span finite subspaces and estimate the corresponding operators on these subspaces. For this estimation we adopt the two-stage deep feature procedure \citep{Xu2021}, in which two variables $\rvx$ and $\rvz$ are related through a nontrivial conditional distribution $P(\rvx \mid \rvz)$, and estimation proceeds by first recovering the conditional expectation of features of $\rvx$ given $\rvz$, and then fitting a target function on the predicted features.



Let $\psi_{\theta_X}:\mathbb{X}\rightarrow \R^{d_1}$ and $\phi_{\theta_Z}:\mathbb{Z}\rightarrow \R^{d_2}$ denote neural feature maps, parameterized by $\theta_X$ and $\theta_Z$, for a random variable $\rvx$ taking the value $ \vx \in \mathbb{X}$ and $\rvz$ taking the value $\vz \in \mathbb{Z}$, and let $y \in \R$ be a scalar outcome. We model the structural function as $f(\vx) := r_\xi \left(\psi_{\theta_X}(\vx)\right) $ using preimage map $r_\xi\,:\R^{d_1}\rightarrow \R$ parameterized by $\xi$. Then the objective is
\begin{align}
    \underset{\xi, \,\theta_X, \,\theta_Z}{\mathrm{min}} \E \left[ \left(y - r_\xi \left( \E[\psi_{\theta_X} (\rvx) \,|\, \rvz ] \right) \right)^2\right].
\end{align}
This casts learning as estimating a conditional covariance operator
and then fitting a preimage model on its prediction. In \cite{Xu2021}, $r_\xi$ is restricted to be linear, in which case the second stage admits a closed-form ridge solution.
The first stage estimates $ \E[\psi_{\theta_X} (\rvx) \,|\, \rvz ]$ in the span of $\phi_{\theta_Z}$ by optimizing the linear map $\mV$ and the feature parameters $\theta_Z$, while keeping $\theta_X$ fixed:
\begin{align}
    \underset{\mV,\; \theta_Z}{\mathrm{min}} \frac{1}{N} \sum_{i=1}^N \|\psi_{\theta_X}(\vx_i) - \mV \phi_{\theta_Z}(\vz_i)\|^2_2 + \lambda_1 \|\mV\|_F^2, 
    \label{eq:df-stage1}
\end{align}
where $\lambda_1 > 0$ is the regularization term. For any fixed $\theta_Z$, the minimizer in $\mV$ has the closed form $\hat{\mV}(\theta_X, \theta_Z) = \Psi \Phi^\top (\Phi \Phi^\top + \lambda_1 \mI)^{-1}$; where $\Psi \in \R^{d_1 \times N}$ and $\Phi \in \R^{d_2 \times N}$ are the feature matrices whose $i$-th columns are $\psi_{\theta_X}(\vx_i)$ and $\phi_{\theta_Z}(\vz_i)$, respectively, and $\mI$ is the identity matrix. The parameters $\theta_Z$ are then updated by gradient descent on the loss evaluated at $\mV = \hat{\mV}(\theta_X, \theta_Z)$, treating the closed-form solution as a differentiable function of $\phi_{\theta_Z}$.
The linear map $\hat{\mV}$ thus obtained serves as a finite-dimensional estimate of the conditional covariance operator on the subspaces spanned by the neural feature maps.


The second stage fits the structural function on the predicted conditional mean features. Given $\hat{\mV}$ from Stage~1 with $\hat{\theta}_Z$ now fixed, we learn $r_\xi$ and refine $\theta_X$ by minimizing:
\begin{align}
    \underset{\xi, \theta_X}{\mathrm{min}} \frac{1}{N} \sum_{i=1}^N \left(y_i - r_\xi \left( \hat{\mV} \phi_{\theta_Z} (\vz_i) \right) \right)^2 + \lambda_2 \,\Omega(\xi),
    \label{eq:df-stage2}
\end{align}
where $\lambda_2 > 0$ is also the regularization term, and $\Omega(\cdot)$ denotes a suitable regularizer on $\xi$. The resulting predictor is obtained as $\hat f(\rvx)=r_{\hat{\xi}}(\psi_{\hat{\theta}_X}(\rvx))$. When the feature maps are fixed, Stage~1 is convex and admits a unique closed-form solution, and Stage~2 is a standard supervised learning problem. Furthermore, if the preimage map is restricted to be linear, Stage~2 similarly admits a closed-form solution \citep{Xu2021}.

Joint optimization of all parameters $(\xi, \theta_X, \theta_Z)$ with respect to the final prediction loss can lead to degenerate representations. As observed by \cite{Xu2021}, without isolating the two stages, the model may bypass the conditional expectation structure, allowing $\phi_{\theta_Z}$ and $r_\xi$ to fit a direct mapping from $\rvz$ to $y$ while rendering the intermediate features $\psi_{\theta_X}$ uninformative. To mitigate this issue, the training typically alternates between two stages. In Stage 1, $\theta_X$ is fixed, ensuring that $\hat{\mV}$ properly estimates the conditional expectation $\E[\psi_{\theta_X}(\rvx) \mid \rvz]$ rather than propagating the outcome gradient directly to $\theta_Z$. In Stage 2, $\theta_Z$ is fixed, and the prediction network is optimized on the projected features. This alternating scheme stabilizes the representation learning. 

\section{Deep Spectral Learning}
\label{sec:Op-bsd-SSM}

\subsection{State Space Model in Operator Form}
\label{subsec:ssm-in-op}

\begin{figure}[t]
    \centering
    \includegraphics[width=.95\linewidth]{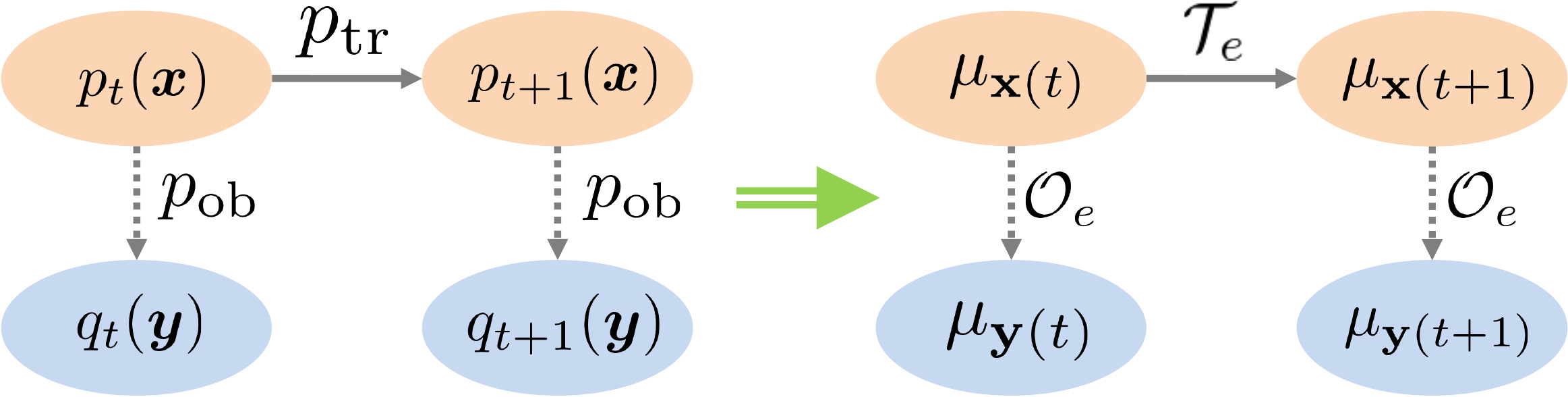}
    \caption{Overview of SSM in Operator Form}
    \label{fig:ssm}
\end{figure}

Section~\ref{subsec:embed-CD} introduced the mean embedding and the conditional covariance operator that provide linear representations of expectations and conditional expectations in Hilbert feature spaces. We apply this to a latent state space model (SSM) for a discrete-time stochastic dynamical system and relate a distribution-level description to an operator-level one.

Let $\{\rvx(t)\}_{t\in\mathbb{T}}$ be a latent stationary and ergodic Markov process on a state space $\mathbb{X}$, and $\{\rvy(t)\}_{t\in\mathbb{T}}$ be an observation process on an observation space $\mathbb{Y}$. The latent dynamics are characterized by a Markov transition kernel from $\rvx(t)$ to $\rvx(t+1)$. When this conditional law admits a density with respect to a reference measure on $\mathbb{X}$, we denote it by $p_{\mathrm{tr}}(\vx_{t+1}\mid \rvx(t)=\vx_t)$. The observation mechanism is characterized by the conditional law of $\rvy(t)$ given $\rvx(t)$. When this conditional law admits a density with respect to a reference measure on $\mathbb{Y}$, we denote it by $p_{\mathrm{ob}}(\vy_t\mid \rvx(t)=\vx_t)$. We do not assume that these densities are known, and we do not model them parametrically. They are introduced to describe how the latent dynamics and the observation channel induce linear maps on probability laws.

At the level of probability densities, $p_{\mathrm{tr}}$ induces 
an evolution of state distributions. If $p_t$ denotes the density of $\rvx(t)$, then the density at the next step satisfies
\begin{align}
p_{t+1}(\vx) \;=\; \int_{\mathbb{X}} p_{\mathrm{tr}}(\vx \mid \rvx(t) =\vz)\, p_t(\vz)\, d\vz.
\end{align}
Similarly, $p_{\mathrm{ob}}$ induces a 
map from state distributions to observation ones. If $q_t$ denotes the density of $\rvy(t)$, then
\begin{align}
q_t(\vy) \;=\; \int_{\mathbb{X}} p_{\mathrm{ob}}(\vy \mid \rvx(t) = \vx)\, p_t(\vx)\, d\vx .
\end{align}
The left side of Figure~\ref{fig:ssm} summarizes these distribution-level transitions through $p_{\mathrm{tr}}$ and $p_{\mathrm{ob}}$.

On the other hand, we now express the same model at the level of embedded distributions. We represent the marginal laws of $\rvx(t)$ and $\rvy(t)$ by their mean embeddings $\mu_{\rvx(t)} \in \mathbb H$ and $\mu_{\rvy(t)} \in \mathbb G$. The conditional covariance operators provide linear maps that propagate these embedded laws.
In particular, we denote by $\mathcal{T}_e$ and $\mathcal{O}_e$ the operators associated with the one-step latent transition and the observation channel, respectively:
\begin{align}
    \mathcal{T}_e := \mathcal{C}_{\rvx(t+1)\mid\rvx(t)},\quad  \mathcal{O}_e := \mathcal{C}_{\rvy(t)\mid\rvx(t)}.
\end{align}
This yields an operator-level state-space representation in which the mean embeddings evolve linearly, as $\mu_{\rvx(t+1)} = \mathcal T_{{e}}\,\mu_{\rvx(t)}$, and $\mu_{\rvy(t)} = \mathcal O_{{e}}\,\mu_{\rvx(t)}$.
The right side of Figure~\ref{fig:ssm} depicts this operator form and its correspondence to the distribution-level description. In \citep{nry2025elto}, $\mathcal{T}_e$ and $\mathcal{O}_e$ are termed the embedded latent transfer operator (ELTO) and embedded observation operator, respectively, and a kernel-based instance of this formulation is proposed. The DSL construction below keeps this operator-level state-space view, but realizes the predictive state coordinates and operator regressions in explicit finite-dimensional neural Galerkin spaces rather than through RKHS kernel evaluations over samples.

\subsection{Stochastic Realization with Deep Features}
\label{subsec:SRwDF}

We adopt the framework of stochastic realization to construct latent state variables directly from observed data. Classical stochastic realization methods, such as balanced stochastic realization based on CCA between past and future observations \citep{Aka75, Desai1985}, estimate a state-space representation by identifying a subspace that yields optimal least-squares predictions. We build on this idea and generalize it to Hilbert spaces and deep features by employing functional CCA for our setting.



Assume the observation process is zero-mean and square-integrable. Fix a Hilbert space $\mathbb{F}$ and a feature map $\phi : \R^d \rightarrow \mathbb{F}$. Let $\mathbb{T}$ denote the set of time indices of a trajectory. For each $t\in\mathbb{T}$, we define the feature process $u_t = \phi(\vy_t) \in\mathbb{F},$
which we regard as an $\mathbb{F}$-valued random variable. We view $\{u_t\}_{t\in\mathbb{T}}$ as elements of the Hilbert space $L^2(\,\Omega, \mathbb{F}\,)$ of square-integrable $\mathbb{F}$-valued random variables. For any $u,\, v\in L^2(\Omega;\mathbb F)$, the inner product is defined by $\langle u, v \rangle_{L^2(\Omega;\mathbb F)} = \E\big[ \langle u, v \rangle_{\mathbb F} \big],$
where $\langle \cdot, \cdot \rangle_{\mathbb F}$ denotes the inner product in $\mathbb F$. We then define the past and future information as the closed linear spans
\begin{align}
    \mathbb{H}^-&:=\overline{\mathrm{span}}\{u_{t-1},u_{t-2},\dots\}\subset L^2(\Omega;\mathbb F),\\ \mathbb{H}^+&:=\overline{\mathrm{span}}\{u_t,u_{t+1},\dots\}\subset L^2(\Omega;\mathbb F).
\end{align}
For arbitrary linear combinations $v^- = \sum_{k=1}^K\alpha_k u_{t-k}$ and $w^- = \sum_{\ell=1}^L\gamma_\ell u_{t-\ell} \in\mathbb H^-$, the inner product in $L^2(\Omega;\mathbb F)$ can be expressed, under weak stationarity, in terms of the lag-covariance function $\Lambda(r) := \E\big[ \langle u_t, u_{t-r}\rangle_{\mathbb F} \big]$ for any time $t\in\mathbb T$, as
\begin{align}
    \langle v^-, w^-\rangle_{L^2(\Omega;\mathbb F)}
  = \sum_{k=1}^K\sum_{\ell = 1}^L\alpha_k\gamma_\ell\,\Lambda(\ell-k).
\end{align}
An analogous expression holds for $\mathbb H^+$ and for past–future cross terms.

For the feature process $u_t \in \mathbb{F}$, define the lag-$r$ covariance operator ($r\in\mathbb{Z}$) as
\begin{align}
    \mathcal{C}_r:\mathbb F\to\mathbb F;\quad \mathcal{C}_r\, x:=\mathbb E[\langle u_t,x\rangle_{\mathbb F}\,u_{t+r}],
\end{align}
where $ x\in\mathbb{F}$, and $\mathcal{C}_r$ is the unique bounded linear operator satisfying $\langle\mathcal{C}_r \,x ,\, h\rangle_\mathbb{F}= \E \left[\langle u_t, \, x\rangle_\mathbb{F} \langle u_{t+r},\, h\rangle_\mathbb{F}\right]$ for all $h\in\mathbb{F}$. Equivalently,  $\mathcal{C}_r$ is denoted as $\E[u_{t+r}\otimes u_t]$. We can define the following vectors as the past and future information at $t$:
\begin{align}
    \vp_t = [u_{t-1},\, u_{t-2},\dots]^\top, \,\, \vf_t = [u_{t},\, u_{t+1},\dots]^\top.
\end{align}
Then the covariance for past process $\E[\vp_t\otimes\vp_t]$ and for future process $\E[\vf_t \otimes \vf_t]$ are considered as block Toeplitz operators; the $(i,\,j)$-block of past is $\mathcal{C}_{j-i}$, and that of  future is $\mathcal{C}_{i-j}$. Also, the covariance between future and past $\E[\vf_t \otimes \vp_t]$ are considered as the block Hankel operator; the  $(i,\,j)$-block is $\mathcal{C}_{i+j-1}$.

The orthogonal projection of $\mathbb{H}^+$ onto $\mathbb{H}^-$ defines the projected space  $\mathbb X_t :=\mathcal P(\mathbb{H}^+ \mid \mathbb{H}^-)$, consisting of all minimum-variance linear predictions of future features based on the past. Because the block Toeplitz covariance operators admit Cholesky factorizations \citep{Chui1982}, one can whiten the past and future coordinates, and the projection of $\mathbb H^+$ onto $\mathbb H^-$ in the whitened basis factors through the block Hankel operator. When the block Hankel operator has finite rank $r$, the image of this projection is therefore $r$-dimensional, and $\mathbb X_t$ is the minimal splitting subspace between $\mathbb H^-$ and $\mathbb H^+$: any subspace between $\mathbb D\subset \mathbb H^-$ satisfying $\mathcal{P}(\mathbb{H}^+ \mid \mathbb{D})= \mathcal{P}(\mathbb{H}^+ \mid \mathbb{H}^-)$ contains $\mathbb{X}_t$. In our setting, the covariance operators on $\mathbb{H}^-$ and $\mathbb{H}^+$ inherit block Toeplitz and Hankel structure from weak stationarity of the feature process $u_t$ as established above, so the results of \citep{LindquistPicci1991, katayama2005subspace} apply directly. A basis for $\mathbb{X}_t$ therefore provides the minimal information from the past needed to predict the future optimally in the least-squares sense, and can serve as latent state coordinates for a Markov model of $u_t$.

To find such a basis we employ CCA between $\mathbb{H}^-$ and $\mathbb{H}^+$ \citep{Desai1985}. That is, we consider the following functional CCA:
\begin{align}
    \underset{v^-\in \mathbb{H}^- \backslash\,\{0\},\,v^+\in\mathbb{H}^+ \backslash\, \{0\}}{\mathrm{max}}\,
\frac{\mathrm{Cov}(v^-,v^+)}{\sqrt{\mathrm{Var}(v^-)}\,\sqrt{\mathrm{Var}(v^+)}},
\end{align}
where for $v^- = \sum_{i\geq1}\alpha_i u_{t-i}$ and $v^+ = \sum_{j\geq 0}\beta_j u_{t+j}$, $\mathrm{Cov}(v^-,v^+)=\sum_{i, j}\,\alpha_i\beta_j\E [\langle u_{t-i},\, u_{t+j}\rangle_{\mathbb{F}}]$, $\mathrm{Var}(v^-)= \mathrm{Cov}(v^-,v^-)$ and $\mathrm{Var}(v^+)= \mathrm{Cov}(v^+,v^+)$. Under weak stationarity, these reduce to Hankel and Toeplitz block operators via $\Lambda(r) = \E[\langle u_t, u_{t-r}\rangle_\mathbb{F}]$ for all time $t$.
The top $r$ canonical directions from this CCA provide an orthonormal basis for the prediction space $\mathbb{X}_t$ \citep{EubankHsing2008, katayama2005subspace}. Specifically, if $\sum_{i \geq 1} \alpha_{l,i}\, u_{t-i}$ denotes the $l$-th canonical direction for $\mathbb{H}^-$ with canonical correlation $s_l$, then the coordinates $x_l(t) := s_l^{1/2} \sum_{i \geq 1} \alpha_{l,i}\, u_{t-i}$ ($l = 1, \ldots, r$) form an orthonormal basis of $\mathbb{X}_t$. Collecting them into $\rvx(t) := [x_1(t), \ldots, x_r(t)]^\top \in \mathbb{R}^r$, one can characterize the one-step evolution of these state coordinates: since the block Hankel operator in our setting admits the same rank-$r$ factorization and block-row-shift property, the construction in \citet{katayama2005subspace, nry2025elto} determines a unique transition matrix $\mA \in \mathbb{R}^{r \times r}$ on the state coordinates as
\begin{equation}
    \label{eq:markov-rep}
    \rvx(t+1) = \mA\,\rvx(t) + \rvw(t),
\end{equation}
where $\rvw(t) := \rvx(t+1) - \mathcal{P}(\rvx(t+1) \mid \rvx(t))$ is the innovation noise satisfying $\rvw(t) \perp \mathbb{H}^-$. This Markov representation gives the optimal minimum-variance evolution of the latent state along time.

In practice, the infinite past and future are truncated to a finite window of length $\ell$, and the population functional CCA is approximated in finite-dimensional Galerkin subspaces. Given a finite number of samples, we define the past and future delay vectors as $\tilde\vp_t := (\vy_{t-1}^\top,\dots,\vy_{t-\ell}^\top)^\top$ and $\tilde\vf_t := (\vy_t^\top,\dots,\vy_{t+\ell-1}^\top)^\top$, and denote by $I_{\mathrm{SR}} \subset \mathbb{T}$ the time index set for which both windows are well-defined. By applying a block feature map $\tilde\phi$ and centering the outputs, we arrange them into feature matrices $\Phi_p$ and $\Phi_f$ over $t \in I_{\mathrm{SR}}$.

Weak stationarity is used to express the past and future covariance blocks through lag-covariances, giving the Toeplitz and Hankel structures used in the CCA realization. The finite-rank condition gives the exact case in which the predictive state space is $r$-dimensional; in finite samples, retaining the top $r$ CCA directions corresponds to an approximate low-rank past--future predictive structure. The function-space assumptions are implemented through the learned block-feature space and the dictionary spaces used for subsequent operator regression.
With the corresponding sample covariance blocks $\mC_-$ and $\mC_+$, as well as the cross-covariance $\mH$, the CCA problem reduces to a singular value decomposition of the normalized cross-covariance block $\mT := \mC_+^{-1/2}\,\mH\,\mC_-^{-1/2}$.
The singular values and vectors of $\mT$ provide finite-dimensional estimates of the canonical correlations and directions, which are then used to extract the state coordinates $\rvx(t)$. Further details of the data-driven algorithm are provided in Appendix~\ref{subsec:app-sr-algorithm}.

\section{Model Architecture and Learning Algorithm}
\label{sec:prop}

\begin{figure*}[t]
    \centering
    \includegraphics[width=1.0\linewidth]{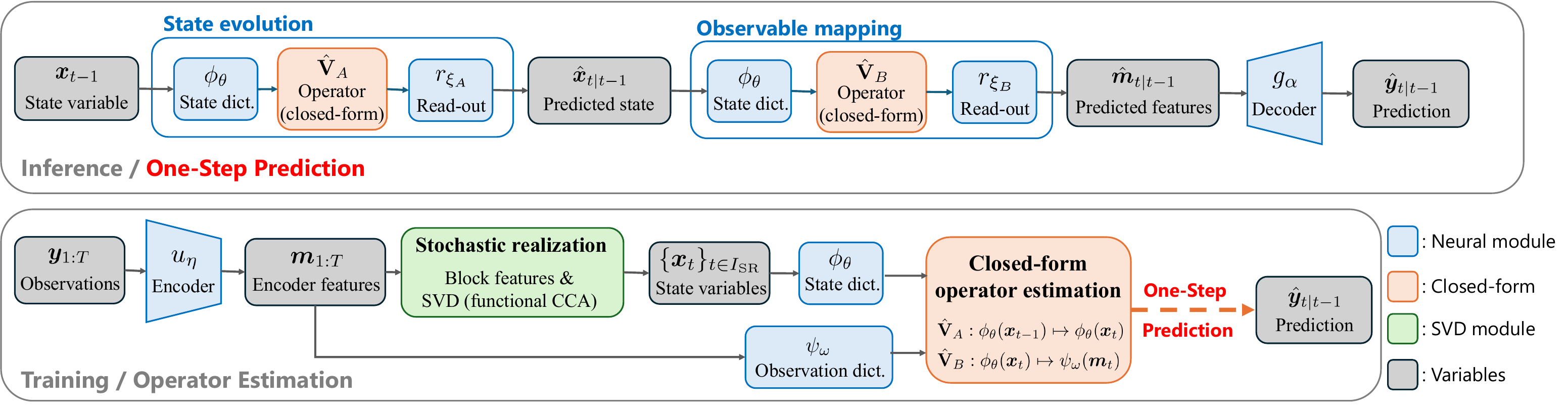}
    \caption{Overall Architecture of Deep Spectral Encoder}
    \label{fig:architecture}
\end{figure*}

\subsection{Design Principles of Deep Spectral Encoder}
\label{subsec:design principles of dse}
In this section, we instantiate the deep spectral learning framework developed in Section~\ref{sec:Op-bsd-SSM} in finite-dimensional deep feature spaces. We introduce the Deep Spectral Encoder, an architecture specifically designed to strictly separate representation learning from dynamics modeling. Jointly optimizing all neural components and operators often leads to degenerate solutions where the encoder absorbs temporal structures, rendering the learned operators trivial. To prevent this, our model adopts a modular design. It sequentially integrates a time-invariant encoder for point-wise feature extraction, a stochastic realization module for state coordinate construction, a closed-form operator estimation step on the feature space, and a time-invariant decoder for observation reconstruction. Figure~\ref{fig:architecture} summarizes the computational graph of this architecture. The detailed empirical algorithms are provided in Appendix~\ref{sec:app-detailed-algorithm}.

\paragraph{Deep Feature Extraction and State Realization:}
\label{subsec:Deep Feature Extraction and State Realization}
To extract predictive latent states from the observation sequence, we first map each observation $\vy_t$ to a feature representation $\vm_t = u_\eta(\vy_t)$ via a time-invariant neural encoder $u_\eta: \mathbb{R}^d \to \mathbb{R}^m$. To construct the past and future block features required for functional CCA without feeding excessively large windows into a single network, we apply a shallow head network $h: \mathbb{R}^\ell \to \mathbb{R}$ to the delay vectors of each feature dimension. Specifically, for the $k$-th dimension of the encoder output, we form the past delay vector $\tilde{\vv}_{k,t} = (u_\eta^{(k)}(\vy_{t-1}), \dots, u_\eta^{(k)}(\vy_{t-\ell}))^\top$ and obtain the scalar block feature $\tilde{\phi}^{(k)}(\tilde{\vp}_t) := h(\tilde{\vv}_{k,t})$. Stacking these scalars yields the past feature $\tilde{\phi}(\tilde{\vp}_t) \in \mathbb{R}^m$, and the future feature $\tilde{\phi}(\tilde{\vf}_t)$ is constructed analogously.

Given these block features for $t \in I_{\mathrm{SR}}$, we directly apply the stochastic realization procedure described in Section~\ref{subsec:SRwDF}. By computing the SVD of the normalized cross-covariance $\mT$, we obtain the top $r$ canonical directions. These directions yield the $r$-dimensional latent state coordinates $\vx_t \in \mathbb{R}^r$, providing a finite-dimensional Markovian representation of the sequence.

\paragraph{Closed-Form Operator Estimation:}
\label{subsec:Closed-Form Operator Estimation}
On the estimated latent state space, we introduce a neural dictionary $\phi_\theta: \mathbb{R}^r \to \mathbb{R}^{d_A}$, and on the observation feature space, another dictionary $\psi_\omega: \mathbb{R}^m \to \mathbb{R}^{d_B}$. These dictionaries induce finite-dimensional subspaces on which we approximate the transfer operator by $\mV_A: \mathbb{R}^{d_A} \to \mathbb{R}^{d_A}$ and the observation operator by $\mV_B: \mathbb{R}^{d_A} \to \mathbb{R}^{d_B}$. 

Let $\Phi_- = [\phi_\theta(\vx_{t-1})]_{t \in I_{\mathrm{SR}}}$, $\Phi_+ = [\phi_\theta(\vx_t)]_{t \in I_{\mathrm{SR}}}$, and $\Psi = [\psi_\omega(\vm_t)]_{t \in I_{\mathrm{SR}}}$ be the feature matrices. Following the deep-feature two-stage procedure, we estimate the operators as ridge-regularized closed-form solutions:
\begin{align}
    \hat{\mV}_A &= \Phi_+ \Phi_-^\top (\Phi_- \Phi_-^\top + \lambda_A \mI)^{-1}, \label{eq:cf-va} \\
    \hat{\mV}_B &= \Psi \widehat{\Phi}_+^\top (\widehat{\Phi}_+ \widehat{\Phi}_+^\top + \lambda_B \mI)^{-1}, \label{eq:cf-vb}
\end{align}
where $\widehat{\Phi}_+ := [\phi_\theta(\hat{\vx}_{t|t-1})]_{t \in I_{\mathrm{SR}}}$ is the matrix of one-step predicted state features computed via the state read-out map $\hat{\vx}_{t|t-1} := r_{\xi_A}(\hat{\mV}_A \phi_\theta(\vx_{t-1}))$. These closed-form solutions coincide with the Galerkin projections of the underlying conditional covariance operators $\mathcal{T}_e$ and $\mathcal{O}_e$, enabling stable estimation without constructing sample-dependent Gram matrices. The read-out maps $r_{\xi_A}$ and $r_{\xi_B}$ are then learned by minimizing the regularized reconstruction errors on the operator-projected features $\hat{\mV}_A \Phi_-$ and $\hat{\mV}_B \widehat{\Phi}_+$. This step corresponds to the second stage of the deep feature procedure, taking the form of the objective in Eq.~\eqref{eq:df-stage2}.

\subsection{Staged Training Strategy}
\label{subsec:Staged Training Strategy}
To successfully train the architecture and enforce the modular design described in Section~\ref{subsec:design principles of dse}, we employ a staged training schedule. At the outset, we freeze the parameters of the encoder and the decoder, and we optimize the dictionaries and read-out maps. The computational cost and the differentiable CCA/SVD implementation are summarized in Appendix~\ref{subsec:app-computational-cost}. This step alternates between closed-form updates of the operators $\hat{\mV}_A$ and $\hat{\mV}_B$ and gradient-based updates of the neural parameters for the dictionaries and read-out maps.

Once the operators and read-outs have stabilized, we unfreeze the encoder and decoder to perform end-to-end refinement along the test-time causal path. We compute the one-step-ahead prediction $\hat{\vy}_{t|t-1} = g_\alpha(r_{\xi_B}(\hat{\mV}_B \phi_\theta(\hat{\vx}_{t|t-1})))$ and optimize the stepwise reconstruction loss $\sum_{t \in I_{\mathrm{SR}}} \|\vy_t - \hat{\vy}_{t|t-1}\|^2$, augmented with a regularization term that maximizes the sum of the canonical correlations obtained from the stochastic realization. This staged schedule separates representation learning from operator estimation and mitigates degenerate solutions observed in joint end-to-end training.

\section{Sequential State Estimation}
\label{sec:application:seq}

Sequential state estimation, often formulated as sequential Bayesian filtering, propagates information about a latent state in a dynamical system from sequentially observed data. Classical Kalman filtering realizes this recursion through posterior mean and covariance updates, alternating between a prediction step that propagates the belief state in time and an innovation step that incorporates new observations \citep{Kal60, Anderson1979}. The idea of expressing Kalman-type updates at the level of operators on feature spaces has been explored in the kernel Kalman rule proposed by \citep{fukumizu13a, GKN19}. In this section we follow this operator-theoretic viewpoint in a general Hilbert-space setting and then describe its finite-dimensional implementation in the feature space learned by DSE.

At time $t$, denote the mean embedding and covariance operator for the prior by $\mu_{\rvx(t)}^-$ and $\mathcal C_{\rvx(t)}^-$, respectively. Let $\mathcal{R}\in\mathcal{L}(\mathbb G)$ and $\mathcal{Q}\in\mathcal{L}(\mathbb{H})$ be self-adjoint, positive semi-definite, bounded noise covariance operators at time $t$.

Then the innovation Kalman update in Hilbert space is
\begin{align}
\mu_{\rvx(t)}^+ &= \mu_{\rvx(t)}^- + \mathcal K_t\,(\psi_y(\vy_t) - \mathcal C_{\rvy(t)\mid\rvx(t)}\mu_{\rvx(t)}^- ), \\
\mathcal C_{\rvx(t)}^+ &= \mathcal C_{\rvx(t)}^- - \mathcal K_t\,\mathcal C_{\rvy(t)\mid\rvx(t)}\,\mathcal C_{\rvx(t)}^-, 
\end{align}
where $\mathcal K_t = \mathcal C_{\rvx(t)}^-\,\mathcal C_{\rvy(t)\mid\rvx(t)}^*\,\bigl(\mathcal C_{\rvy(t)\mid\rvx(t)}\mathcal C_{\rvx(t)}^-\mathcal C_{\rvy(t)\mid\rvx(t)}^* + \mathcal R\bigr)^{-1}$, and $\psi_y(\vy_t)\in\mathbb{G}$ is the feature map of the observation. These are the operator‑level Kalman updates.

For the time update of state process, we employ the conditional covariance operator for state process $\mathcal{C}_{\rvx({t+1})|\rvx(t)}$, and its Kalman update formulated as
\begin{align}
    \mu_{\rvx(t+1)}^- &= \mathcal C_{\rvx(t+1)|\rvx(t)}  \mu_{\rvx(t)}^+, \\
    \mathcal{C}_{\rvx(t+1)}^- &= \mathcal C_{\rvx(t+1)|\rvx(t)}\mathcal{C}_{\rvx(t)}^+\mathcal C_{\rvx(t+1)|\rvx(t)}^* + \mathcal{Q}.
\end{align}
Applying these updates sequentially yields an operator-level forward filter for the latent state.

In practice, the operators $\mathcal{C}_{\rvx(t+1)\mid\rvx(t)}$ and $\mathcal{C}_{\rvy(t)\mid\rvx(t)}$ are represented by the matrices $\hat{\mV}_A$ and $\hat{\mV}_B$ estimated in Section~\ref{subsec:design principles of dse}. The process noise covariance $\mathcal{Q}$ and observation noise covariance $\mathcal{R}$ are estimated from the empirical residuals of these operators on the training trajectory. Then the filtering recursion is performed entirely in the feature space by propagating and updating the moments through the learned operators without explicit Gram matrix constructions. This feature-space recursion corresponds to the operator-level Kalman-type update proposed in \citet{GKN19}. The detailed estimation of noise covariances and the recursive filtering algorithm are provided in Appendix~\ref{subsec:app-filtering-algorithm}. Finally, the latent state is recovered via the read-out map as $\hat{\vx}_t = r_{\hat{\xi}_A}(\bm\mu_t^+)$.

\begin{table}[t]
  \centering
  \caption{Quad-Link Pendulum Images}
  \label{tab:quadlink}
  \small
  \begin{tabular}{llcc}
    \toprule
    Method & length & Without noise & With noise \\
    \midrule
    RKN      & 1.5k & 0.2198 $\pm$ 0.0132 & 0.2944 $\pm$ 0.0236 \\
    LSTM     & 1.5k & 0.2527 $\pm$ 0.0152 & 0.3167 $\pm$ 0.0223 \\
    ELTO-KF  & 1.5k & 0.2175 $\pm$ 0.0130 & 0.2942 $\pm$ 0.0215 \\
    \textbf{DSE} (ours) & 1.5k  & \textbf{0.2006} $\pm$ 0.0228 & \textbf{0.2593} $\pm$ 0.0274 \\
    \midrule
    RKN      & 15k  & 0.2838 $\pm$  0.0190 & 0.3258  $\pm$ 0.0241 \\
    LSTM     & 15k  & 0.3050  $\pm$ 0.0183 & 0.3158 $\pm$ 0.0244\\
    ELTO-KF  & 15k  & 0.2924  $\pm$ 0.0175 & 0.2936 $\pm$ 0.0207\\
    \textbf{DSE} (ours) & 15k  & \textbf{0.2615} $\pm$ 0.0211& \textbf{0.2683}  $\pm$ 0.0254\\
    \bottomrule
  \end{tabular}
\end{table}


\paragraph{Numerical Results:} We evaluate on quad-link pendulum image sequences of size $48\times 48$ under a clean regime and a corrupted regime with occlusion and geometric distortion. Initial joint angles are sampled uniformly from $[-\pi, \pi]$ with zero initial velocities, viscous friction is $0.1$, the simulation time step is $10^{-4}$ seconds, frames are recorded every $0.05$ seconds, and the first five frames in each sequence are kept clean. One-step prediction errors are reported as mean squared error for training--test sequence lengths of $1.5\times 10^3$ and $1.5\times 10^4$. For the one-step comparison in Table~\ref{tab:quadlink}, baselines include a Recurrent Kalman Network (RKN) \citep{BPG+19} and an LSTM predictor, as well as ELTO-Kalman filtering (ELTO-KF) \citep{nry2025elto}. We keep the encoder resolution, data splits, preprocessing, simulator and noise parameters fixed across methods, including the corrupted setting with $\rho = 0.2$ and thresholds $[0, 0.25, 0.75, 1.0]$.

Table~\ref{tab:quadlink} reports the mean and standard deviation of the one-step prediction MSE for all methods under the clean and corrupted regimes. Across all training lengths and noise settings, DSE attains the lowest average error and consistently improves upon the RKN and the LSTM, while also outperforming the ELTO-based predictor.

\begin{table*}[t]
  \centering
  \caption{Multi-step prediction on Quad-Link benchmark}
  \label{tab:quadlink-multistep}
  \small
  \begin{tabular}{lccccc}
    \toprule
    Method & $H=1$ & $H=5$ & $H=10$ & $H=20$ & $H=50$ \\
    \midrule
    ELTO-KF & 0.2942 $\pm$ 0.0215 & 0.3104 $\pm$ 0.0256 & 0.3309 $\pm$ 0.0304 & 0.3068 $\pm$ 0.0403 & 0.3953 $\pm$ 0.0503 \\
    RKN     & 0.2944 $\pm$ 0.0236 & 0.3209 $\pm$ 0.0308 & 0.3452 $\pm$ 0.0408 & 0.3758 $\pm$ 0.0501 & 0.4057 $\pm$ 0.0607 \\
    LSTM    & 0.3167 $\pm$ 0.0233 & 0.3557 $\pm$ 0.0350 & 0.3852 $\pm$ 0.0451 & 0.4109 $\pm$ 0.0554 & 0.4307 $\pm$ 0.0658 \\
    NCDSSM  & 0.2804 $\pm$ 0.0358 & 0.3005 $\pm$ 0.0406 & 0.3259 $\pm$ 0.0457 & 0.3607 $\pm$ 0.0555 & 0.4004 $\pm$ 0.0702 \\
    \textbf{DSE} (ours) & \textbf{0.2593} $\pm$ 0.0274 & \textbf{0.2754} $\pm$ 0.0308 & \textbf{0.2952} $\pm$ 0.0357 & \textbf{0.3256} $\pm$ 0.0456 & \textbf{0.3805} $\pm$ 0.0551 \\
    \bottomrule
  \end{tabular}
\end{table*}

\paragraph{Results on Multi-Step Prediction:}
To evaluate rollout stability beyond one-step prediction, we further consider multi-step prediction on the $1.5\times 10^3$ corrupted quad-link benchmark. At each time $t$, each method is first conditioned on the observations up to $t$ using its own state-update mechanism and then predicts $\hat{\vy}_{t+H\mid t}$ for $H\in\{1,5,10,20,50\}$ without using observations after $t$. We also include Neural Continuous-Discrete State Space Model (NCDSSM) \citep{ansari2023ncdssm} as an additional neural latent state-space baseline, adapting its image encoder and decoder to the $48\times48$ grayscale observations. Table~\ref{tab:quadlink-multistep} reports the resulting prediction MSEs. In this setting, DSE attains the lowest mean MSE across the evaluated horizons, and the multi-step results complement the one-step comparison by measuring how prediction errors compound after the filtering update is no longer conditioned on new observations.



\section{Mode Decomposition}
\label{sec:application:mode}

Operator-theoretic analysis describes nonlinear dynamics via linear operators acting on observables, which enables spectral decompositions and mode interpretations \citep{brunton2022}. Let $\mathcal{K}$ denote the Koopman operator and let $\{(\lambda_i,{\varphi}_i)\}_{i\ge 1}$ be its eigenpairs. For an observable $f:\mathbb X\to\mathbb C^d$ that lies in the span of these eigenfunctions, the $t$-step conditional expectation admits the expansion
\begin{align}
    (\mathcal K^{t} f)(\vx) = \sum_{i} \lambda_i^{t}\, \varphi_i(\vx)\, \vv_i,
\end{align}
where $\vv_i$ are Koopman modes and each mode evolves with a single complex rate. This Koopman mode decomposition separates oscillatory frequencies from growth or decay, and motivates finite-dimensional approximations such as dynamic mode decomposition (DMD) \citep{sch2010}.

We analyze the learned dynamics through the Koopman operator associated with the latent Markov process in Section~\ref{subsec:ssm-in-op}. For an observable function $f:\mathbb X \to \mathbb C^d$, the Koopman operator $\mathcal K$ maps $f$ to its conditional expectation under the transition density $p_{\mathrm{tr}}$, defined as \citep{Klus2019}:
\begin{align}
\hspace{-2.5mm}
(\mathcal K f)(\vx_t)
&= \!\int_{\mathbb X} p_{\mathrm{tr}}(\vx_{t+1}\mid \rvx(t) = \vx_t)\, f(\vx_{t+1})\, d\vx_{t+1} \\
&= \E\!\left[f\!\left(\rvx(t+1)\right)\mid \rvx(t)=\vx_t\right].
\end{align}
For the observation model $p_{\mathrm{ob}}$ defined in Section~\ref{subsec:ssm-in-op}, we consider the conditional mean observation function $f_{\mathrm{ob}}$ given by
\begin{align}
f_{\mathrm{ob}}(\vx_t)
&= \int_{\mathbb Y} p_{\mathrm{ob}}(\vy_t \mid \rvx(t)=\vx_t)\, \vy_t\, d\vy_t,
\end{align}
so that
$(\mathcal K f_{\mathrm{ob}})(\vx_t) = \mathbb E[\rvy(t+1)\mid \rvx(t)=\vx_t]$.

We define observables in the Hilbert space $\mathbb H$ and write $\mathcal K_{\mathbb H}$ for the Koopman operator projected onto $\mathbb H$. Under the assumption that $\mathbb{E}[f_{\mathrm{ob}}(\rvx(t+1))\mid \rvx(t)=\cdot]\in \mathbb H$, the projected operator satisfies the identity \citep{Fukumizu2004, nry2025elto}:
\begin{align}
\mathcal C_{\rvx(t)}\, (\mathcal K_{\mathbb H} f_{\mathrm{ob}}) = \mathcal C_{\rvx(t)\rvx(t+1)}\, f_{\mathrm{ob}}.
\end{align}
Since the inverse of $\mathcal C_{\rvx(t)}$ may be ill-posed, we work with the ridge-regularized projected Koopman operator
$$
\mathcal K_{\mathbb H,\lambda} = (\mathcal C_{\rvx(t)}+\lambda \,\mathcal I)^{-1} \mathcal C_{\rvx(t)\rvx(t+1)},
$$
which can take the adjoint and yield
\begin{align}
\mathcal K_{\mathbb H,\lambda}^{*}
&= \mathcal C_{\rvx(t+1)\rvx(t)}\, (\,\mathcal C_{\rvx(t)}+\lambda\, \mathcal I\,)^{-1}
= \mathcal T_e,
\end{align}
where $\mathcal T_e$ is the conditional covariance operator defined in Section~\ref{subsec:ssm-in-op} that propagates embedded state laws forward in time. As discussed in Section~\ref{sec:prop}, $\mV_A$ is the matrix representation of the Galerkin projection of $\mathcal T_e$ onto the subspace spanned by the learned feature dictionaries and is estimated by the closed-form normal equations in Eq.~\eqref{eq:cf-va}. This adjoint relation connects Koopman spectral analysis to the learned transfer operator.

To compute the spectrum and modes, we perform eigen-decomposition induced by learned $\hat\mV_A$.
When the sampling interval $\Delta t$ is known, we convert to continuous-time eigenvalues via
$
\mu_i = \Delta t^{-1} \log \lambda_i,
$
so that the real part of $\mu_i$ gives the decay rate and imaginary part gives the angular frequency.
In the numerical experiments below, we quantify spectrum recovery by comparing the estimated eigenvalues to a reference spectrum computed from the ground-truth dynamics, and we compare the results with several baseline methods described later.

\begin{figure}[t]
    \centering
    \includegraphics[width=1.05\linewidth]{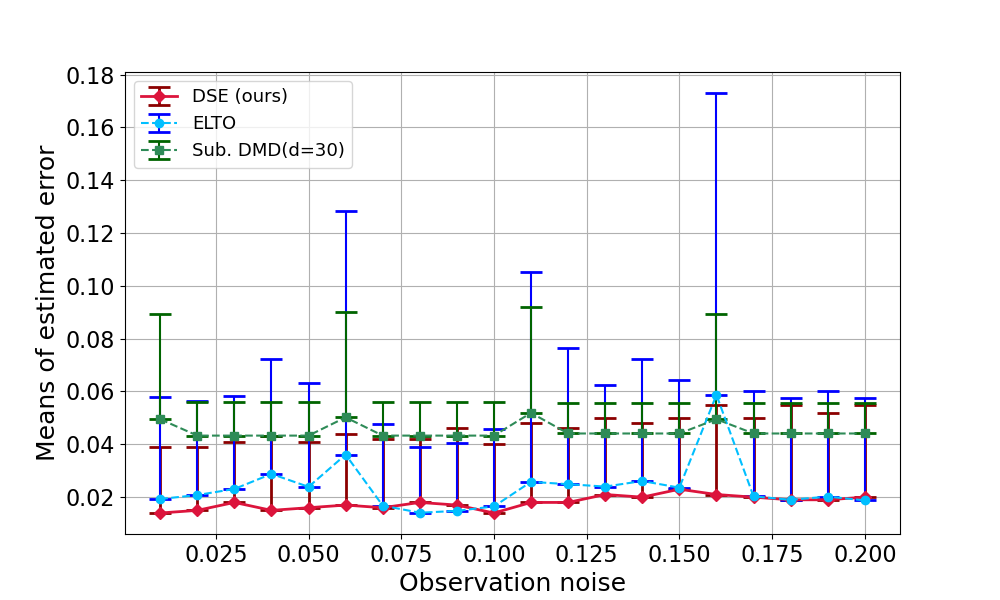}
    \caption{Results for VDP Oscillator}
    \label{fig:exp-vdp-zoom}
\end{figure}

\begin{table}[t]
\centering
\caption{Results at Representative Noise Levels (VDP)}
\label{tab:kmd-vdp}
\setlength{\tabcolsep}{3pt}
\resizebox{\columnwidth}{!}{
\begin{tabular}{c|cccc}
\hline
noise & DSE (ours) & ELTO & sDMD & hDMD \\
\hline
0.05 & 0.016$\pm$0.025 & 0.024$\pm$0.039 & 0.043$\pm$0.013 & 0.076$\pm$0.042 \\
0.10 & 0.014$\pm$0.026 & 0.016$\pm$0.029 & 0.043$\pm$0.013 & 0.109$\pm$0.072 \\
0.15 & 0.023$\pm$0.028 & 0.024$\pm$0.041 & 0.044$\pm$0.012 & 0.129$\pm$0.082 \\
0.20 & 0.020$\pm$0.035 & 0.019$\pm$0.039 & 0.044$\pm$0.012 & 0.143$\pm$0.091 \\
\hline
Avg. & \textbf{0.019}$\pm$0.029 & 0.024$\pm$0.046 & 0.045$\pm$0.020 & 0.104$\pm$0.074 \\
\hline
\end{tabular}
}
\end{table}

\begin{figure}[t]
    \centering
    \includegraphics[width=1.1\linewidth]{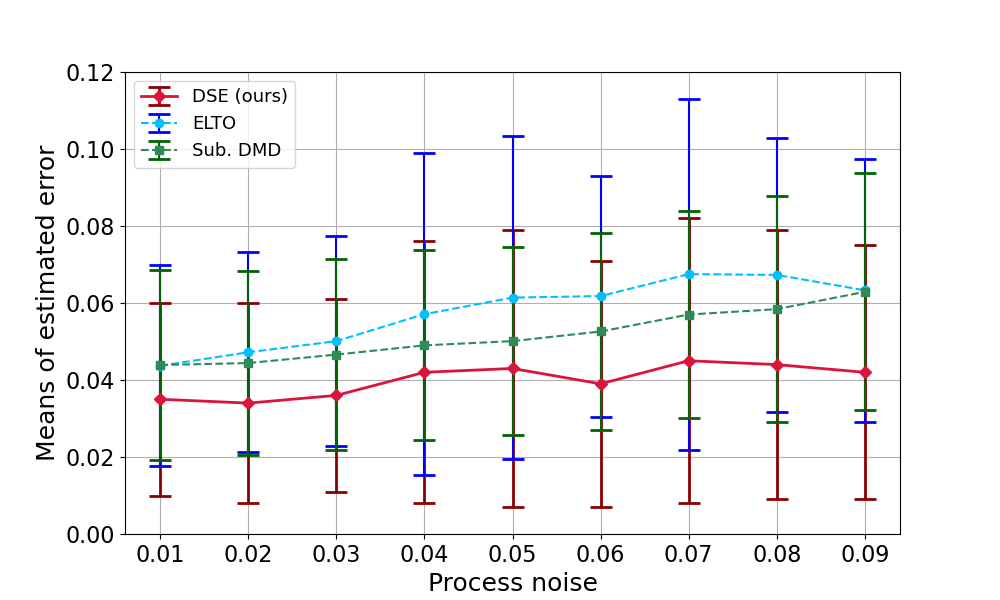}
    \caption{Results for SL Oscillator}
    \label{fig:exp-sl-zoom}
\end{figure}

\begin{table}[t]
\centering
\caption{Results at Representative Noise Levels (SL)}
\label{tab:kmd-sl}
\setlength{\tabcolsep}{3pt}
\resizebox{\columnwidth}{!}{
\begin{tabular}{c|cccc}
\hline
noise  & DSE (ours) & ELTO & sDMD & eDMD \\
\hline
0.01 & 0.035$\pm$0.025 & 0.044$\pm$0.026 & 0.044$\pm$0.025 & 0.325$\pm$0.129  \\
0.05 & 0.043$\pm$0.036 & 0.061$\pm$0.042 & 0.050$\pm$0.024 & 0.318$\pm$0.132  \\
0.09 & 0.042$\pm$0.033 & 0.063$\pm$0.034 & 0.063$\pm$0.031 & 0.323$\pm$0.126  \\
\hline
Avg. & \textbf{0.041}$\pm$0.032  & 0.058$\pm$0.034 & 0.052$\pm$0.026 & 0.321$\pm$0.130  \\
\hline
\end{tabular}
}
\end{table}

\paragraph{Numerical Results:}
To evaluate the robustness of DSE in recovering Koopman spectra under noise, we consider two standard nonlinear oscillators: the Van der Pol (VDP) oscillator and Stuart-Landau (SL) oscillator with observation and process noises.

We first examine the VDP oscillator against zero-mean Gaussian observation noise. By varying the noise variance $\sigma^2_\mathrm{obs}$ between $0.01$ and $0.20$, we generate independent noisy trajectories and compare the estimated eigenvalues with a reference spectrum derived from the noise-free limit cycle. The estimation error is defined as the absolute difference between the estimated and reference eigenvalues averaged over multiple trials. Next, we evaluate the SL oscillator to assess robustness against dynamic process noise. Following the same evaluation protocol, we fix the observation noise variance at $0.01$ and vary the process noise variance $\epsilon_{\mathrm{proc}}$ between $0.01$ and $0.09$. For this system, the estimated eigenvalues are compared with an analytical reference spectrum derived from the asymptotic theory under weak process noise \citep{bag2014}. The explicit governing equations, numerical integration schemes, and procedures for computing the reference spectra for both systems are detailed in Appendix~\ref{subsec:app-vdp} and~\ref{subsec:app-sl}, while their data generation hyperparameters are summarized in Tables~\ref{tab:vdp_setting} and~\ref{tab:sl_setting}.

\paragraph{Baselines:}
DSE is compared with several DMD-based approaches: extended DMD (eDMD) \citep{WKR15}, Hankel DMD (hDMD) \citep{AM17}, subspace DMD (sDMD) \citep{TK21}, and ELTO \citep{nry2025elto}, whose learned transfer operator spectrum serves as the Koopman spectral estimator.

For VDP, hDMD and sDMD are applied directly to noisy scalar observations with length-$30$ Hankel delay coordinates. ELTO and DSE first infer a latent state trajectory via the realization step, then estimate the operator, and report the eigenvalues for the Koopman spectrum. For SL, ELTO and DSE again infer latent states from the observation process.
eDMD and sDMD lift observations using complex exponentials based on the polar Fourier feature map. 

Figures~\ref{fig:exp-vdp-zoom} and~\ref{fig:exp-sl-zoom} plot mean eigenvalue estimation errors with one-standard-deviation bars, and Tables~\ref{tab:kmd-vdp} and~\ref{tab:kmd-sl} report the same metric at representative noise levels; methods with substantially larger errors are omitted for clarity. On VDP, DSE achieves the lowest mean error across noise levels; ELTO is the closest baseline, matching DSE at several levels, but DSE exhibits smaller standard deviations. The DMD baselines incur larger errors, with hDMD degrading notably as noise increases. On SL, DSE again achieves the lowest mean error, with sDMD and ELTO close behind, and remains more stable than ELTO.
 \section{Conclusion}
 \label{sec:cclsn}

This work developed a deep spectral learning framework for stochastic dynamical systems and instantiated it as DSE, an operator-based latent state space model that learns time-invariant encoders and linear transfer and observation operators in finite-dimensional deep feature spaces. Using stochastic realization with covariance and conditional covariance operators, DSE computes predictive state coordinates and ridge-regularized closed-form operator estimates via Galerkin regression, enabling Kalman-style sequential state estimation and Koopman spectral mode decomposition in feature space. Experiments on nonlinear oscillators and image sequences show robust forecasting and spectrum recovery under noise and partial observability, with performance comparable to or better than baselines.


\begin{acknowledgements}
This work was partially supported by JSPS KAKENHI Grant Numbers JP22H05106 and JP26H02498, JST CREST Grant Number JPMJCR1913, and JST SPRING Grant Number JPMJSP2138.
\end{acknowledgements}

\bibliography{uai2026-template}

@InProceedings{nry2025elto,
  title = 	 {Learning Stochastic Nonlinear Dynamics with Embedded Latent Transfer Operators},
  author = {Ke, N. and Tanaka, R. and Kawahara, Y.},
  booktitle = 	 {Proceedings of the 28th International Conference on Artificial Intelligence and Statistics},
  pages = 	 {4861--4869},
  year = 	 {2025},
  volume = 	 {258},
  series = 	 {Proceedings of Machine Learning Research},
  publisher =    {PMLR},
}

@InProceedings{ansari2023ncdssm,
  author = {Ansari, A. F. and Heng, A. and Lim, A. and Soh, H.},
  title = 	 {Neural Continuous-Discrete State Space Models for Irregularly-Sampled Time Series},
  booktitle = 	 {Proceedings of the 40th International Conference on Machine Learning},
  pages = 	 {926--951},
  volume = 	 {202},
  series = 	 {Proceedings of Machine Learning Research},
  month = 	 {23--29 Jul},
  year = 	 {2023},
}

@article{hasan2022,
  author = {Hasan, A. and Pereira, J. M. and Farsiu, S. and Tarokh, V.},
  journal={IEEE Transactions on Signal Processing}, 
  title={Identifying Latent Stochastic Differential Equations}, 
  volume={70},
  pages={89-104},
  year={2022},
}

@InProceedings{duncker19,
  author = {Duncker, L. and Bohner, G. and Boussard, J. and Sahani, M.},
  title = 	 {Learning interpretable continuous-time models of latent stochastic dynamical systems},
  booktitle = 	 {Proceedings of the 36th International Conference on Machine Learning},
  pages = 	 {1726--1734},
  volume = 	 {97},
  series = 	 {Proceedings of Machine Learning Research},
  month = 	 {09--15 Jun},
  year = 	 {2019},
}

@article{Chui1982,
   author = {Chui, C. K. and Ward, J. D. and Smith, P. W.},
   title = {Cholesky factorization of positive definite bi-infinite matrices},
   journal = {Numerical Functional Analysis and Optimization},
   volume = {5},
   number = {1},
   pages = {1--20},
   year = {1982}
}

@article{LindquistPicci1991,
   author = {Lindquist, A. and Picci, G.},
   title = {A geometric approach to modelling and estimation of linear stochastic systems},
   journal = {Journal of Mathematical Systems, Estimation and Control},
   volume = {1},
   number = {3},
   pages = {241--333},
   year = {1991},
   type = {Journal Article}
}

@article{EubankHsing2008,
   author = {Eubank, R. L. and Hsing, T.},
   title = {Canonical correlation for stochastic processes},
   journal = {Stochastic Processes and their Applications},
   volume = {118},
   number = {9},
   pages = {1634--1661},
   year = {2008},
   type = {Journal Article}
}

@article{brunton2022,
  author = {Brunton, S. L. and Budi{\v{s}}i{\'c}, M. and Kaiser, E. and Kutz, J. N.},
  title   = {Modern {K}oopman Theory for Dynamical Systems},
  journal = {SIAM Review},
  year    = {2022},
  volume  = {64},
  number  = {2},
  pages   = {229--340},
}

@inproceedings{SGSS07,
   author = {Smola, A. and Gretton, A. and Song, L. and Sch{\"{o}}lkopf, B.},
   title = {A {H}ilbert Space Embedding for Distributions},
   booktitle = {Algorithmic Learning Theory},
   series    = {Lecture Notes in Computer Science},
   volume    = {4754},
   pages = {13--31},
   year = {2007},
   type = {Conference Proceedings}
}

@inproceedings{Hsu09,
  author = {Hsu, D. J. and Kakade, S. M. and Zhang, T.},
  title        = {A Spectral Algorithm for Learning Hidden {M}arkov Models},
  booktitle    = {22nd Annual Conference on Learning Theory - {COLT} 2009, Montreal, Quebec,
                  Canada, June 18-21},
  year         = {2009},
}

@inproceedings{Kawahara06,
  author = {Kawahara, Y. and Yairi, T. and Machida, K.},
  title        = {A Kernel Subspace Method by Stochastic Realization for Learning Nonlinear
                  Dynamical Systems},
  booktitle    = {Proceedings of the 19th Conference on Neural Information Processing Systems},
  pages        = {665--672},
  publisher    = {{MIT} Press},
  year         = {2006},
}

@article{AM17,
   author = {Arbabi, H. and Mezi{\'c}, I.},
   title = {Ergodic Theory, Dynamic Mode Decomposition, and Computation of Spectral Properties of the {K}oopman Operator},
   journal = {SIAM Journal on Applied Dynamical Systems},
   volume = {16},
   number = {4},
   pages = {2096--2126},
   year = {2017},
   type = {Journal Article}
}

@article{Aka75,
   author = {Akaike, H.},
   title = {Markovian representation of stochastic processes by canonical variables},
   journal = {SIAM Journal on Control},
   volume = {13},
   number = {1},
   pages = {162--173},
   year = {1975},
   type = {Journal Article}
}

@article{fukumizu13a,
  author = {Fukumizu, K. and Song, L. and Gretton, A.},
  title   = {Kernel {B}ayes' Rule: Bayesian Inference with Positive Definite Kernels},
  journal = {Journal of Machine Learning Research},
  year    = {2013},
  volume  = {14},
  number  = {118},
  pages   = {3753--3783},
}

@article{Klus2019,
   title={Eigendecompositions of Transfer Operators in Reproducing Kernel {H}ilbert Spaces},
   volume={30},
   number={1},
   journal={Journal of Nonlinear Science},
   publisher={Springer Science and Business Media LLC},
   author = {Klus, S. and Schuster, I. and Muandet, K.},
   year={2019},
   pages={283--315} 
}

@book{katayama2005subspace,
  author = {Katayama, T.},
  title     = {Subspace Methods for System Identification},
  series    = {Communications and Control Engineering},
  publisher = {Springer Nature},
  year      = {2005},
}

@article{Fukumizu2004,
   author = {Fukumizu, K. and Bach, F. R. and Jordan, M. I.},
   title = {Dimensionality reduction for supervised learning with reproducing kernel {H}ilbert spaces},
   journal = {Journal of Machine Learning Research},
   volume = {5},
   pages = {73--99},
   year = {2004},
   type = {Journal Article}
}

@inproceedings{Sriperumbudur2008,
  author = {Sriperumbudur, B. K. and Gretton, A. and Fukumizu, K. and Lanckriet, G. R. G. and Sch{\"{o}}lkopf, B.},
  title        = {Injective {H}ilbert Space Embeddings of Probability Measures},
  booktitle    = {21st Annual Conference on Learning Theory},
  pages        = {111--122},
  year         = {2008},
}

@inproceedings{Song2009,
author = {Song, L. and Huang, J. and Smola, A. and Fukumizu, K.},
title = {{H}ilbert space embeddings of conditional distributions with applications to dynamical systems},
year = {2009},
booktitle = {Proceedings of the 26th Annual International Conference on Machine Learning},
pages = {961--968},
numpages = {8},
}

@inproceedings{Song2010,
author = {Song, L. and Boots, B. and Siddiqi, S. M. and Gordon, G. and Smola, A.},
title = {{H}ilbert space embeddings of hidden {M}arkov models},
year = {2010},
booktitle = {Proceedings of the 27th International Conference on Machine Learning},
pages = {991--998},
numpages = {8},
}

@article{baker1973,
 author = {Baker, C. R.},
 journal = {Transactions of the American Mathematical Society},
 pages = {273--289},
 publisher = {American Mathematical Society},
 title = {Joint Measures and Cross-Covariance Operators},
 volume = {186},
 year = {1973}
}

@inproceedings{
Xu2021,
title={Learning Deep Features in Instrumental Variable Regression},
author = {Xu, L. and Chen, Y. and Srinivasan, S. and de Freitas, N. and Doucet, A. and Gretton, A.},
booktitle={International Conference on Learning Representations},
year={2021},
}

@article{WKR15,
  title = {A Data-Driven Approximation of the {K}oopman Operator: {E}xtending Dynamic Mode Decomposition},
  author = {Williams, M. O. and Kevrekidis, I. G. and Rowley, C. W.},
  journal = {Journal of Nonlinear Science},
  volume = {25},
  number = {6},
  pages = {1307--1346},
  year = {2015},
}

@book{Hsing2015,
  title     = {Theoretical Foundations of Functional Data Analysis, with an Introduction to Linear Operators},
  author = {Hsing, T. and Eubank, R.},
  year      = {2015},
  publisher = {Wiley},
}

@article{Desai1985,
  author = {Desai, U. B. and Pal, D. and Kirkpatrick, R. D.},
  title    = {A realization approach to stochastic model reduction},
  journal  = {International Journal of Control},
  volume   = {42},
  number   = {4},
  pages    = {821--838},
  year     = {1985},
  publisher= {Taylor \& Francis},
}

@InProceedings{BPG+19,
  title = 	 {Recurrent {K}alman Networks: Factorized Inference in High-Dimensional Deep Feature Spaces},
  author = {Becker, P. and Pandya, H. and Gebhardt, G. and Zhao, C. and Taylor, C. J. and Neumann, G.},
  booktitle = 	 {Proceedings of the 36th International Conference on Machine Learning},
  pages = 	 {544--552},
  year = 	 {2019},
  volume = 	 {97},
  publisher =    {PMLR},
}

@article{GKN19,
   author = {Gebhardt, G. H. W. and Kupcsik, A. and Neumann, G.},
   title = {The kernel {K}alman rule: {E}fficient nonparametric inference by recursive least-squares and subspace projections},
   journal = {Machine Learning},
   volume = {108},
   pages = {2113--2157},
   year = {2019},
   type = {Journal Article}
}

@inproceedings{grunewalder2012,
  author = {Gr{\"{u}}new{\"{a}}lder, S. and Lever, G. and Baldassarre, L. and Patterson, S. and Gretton, A. and Pontil, M.},
  title        = {Conditional mean embeddings as regressors},
  booktitle    = {Proceedings of the 29th International Conference on Machine Learning},
  year         = {2012},
}

@inproceedings{TK21,
   author = {Takeishi, N. and Kawahara, Y.},
   title = {Learning Dynamics Models with Stable Invariant Sets},
   booktitle = {Proceedings of the 35th AAAI Conference on Artificial Intelligence},
   pages = {9782--9790},
   year = {2021},
   type = {Conference Proceedings}
}

@article{bag2014,
    author = {Bagheri, S.},
    title = {Effects of weak noise on oscillating flows: Linking quality factor, Floquet modes, and {K}oopman spectrum},
    journal = {Physics of Fluids},
    volume = {26},
    number = {9},
    pages = {094104},
    year = {2014},
}

@article{Kal60,
   author = {Kalman, R.E.},
   title = {A New Approach to Linear Filtering and Prediction Problems},
   journal = {Trans.\@ of the ASME, Journal of Basic Engineering},
   volume = {82},
   pages = {35--45},
   year = {1960}
}

@book{Anderson1979,
  author = {Anderson, B. D. O. and Moore, J. B.},
  publisher = {Prentice-Hall},
  title = {Optimal Filtering},
  username = {aude.hofleitner},
  year = 1979
}

@book{doucetfg01,
  author = {Doucet, A. and de Freitas, N. and Gordon, N. J.},
  title        = {Sequential {M}onte {C}arlo Methods in Practice},
  series       = {Statistics for Engineering and Information Science},
  publisher    = {Springer},
  year         = {2001},
}

@article{krishnan2015,
  author = {Krishnan, R. G. and Shalit, U. and Sontag, D.},
  title={Deep {K}alman filters},
  journal={arXiv preprint arXiv:1511.05121},
  year={2015}
}

@article{sch2010,
author="Schmid, P. J.",
title="Dynamic mode decomposition of numerical and experimental data",
journal="Journal of Fluid Mechanics",
publisher="Cambridge University Press (CUP)",
year="2010",
volume="656",
pages="5--28",
}

\newpage

\onecolumn

\title{Deep Spectral Learning of Embedded Latent Transfer Operators\\for Stochastic Dynamical Systems (Supplementary Material)}
\maketitle


\appendix

\section{Details of Numerical Algorithms}
\label{sec:app-detailed-algorithm}

This appendix provides the formal pseudo-code for the training and inference procedures of the Deep Spectral Encoder (DSE). The following algorithms instantiate the operator-theoretic framework described in the main text into implementable steps.

\subsection{Empirical Stochastic Realization}
\label{subsec:app-sr-algorithm}

Algorithm~\ref{alg:sr} details the computational procedure for the empirical stochastic realization introduced in Sections~\ref{subsec:SRwDF} and~\ref{sec:prop}. This procedure extracts the finite-dimensional Markovian state coordinates from the encoder features.

\paragraph{Block Feature Construction.}
To capture temporal dependencies over the window length $\ell$ without excessively increasing the parameter space, the sequence of encoder features $\vm_t = u_\eta(\vy_t)$ is rearranged into delay vectors dimension-wise. A shared shallow network $h$ maps these delay vectors to scalar features, which are then concatenated to form the past and future block features $\tilde{\phi}(\tilde{\vp}_t)$ and $\tilde{\phi}(\tilde{\vf}_t)$. 


\paragraph{Subspace Identification.}
The covariance and cross-covariance operators on the Galerkin subspace are empirically estimated as the matrices $\mC_-, \mC_+$, and $\mH$. The functional canonical correlation analysis (CCA) reduces to the singular value decomposition (SVD) of the normalized cross-covariance matrix $\mT$, where diagonal loading is applied before whitening for numerical stability. The top $r$ canonical directions extracted from the right singular vectors identify the predictive coordinates $\vx_t$ for the latent process.

\begin{algorithm}[h]
\caption{Empirical Stochastic Realization via Functional CCA}
\label{alg:sr}
\begin{algorithmic}[1]
\REQUIRE Encoder features $\{\vm_t\}_{t=1}^T \subset \mathbb{R}^m$, window length $\ell$, state dimension $r$, head network $h$.
\ENSURE Latent state coordinates $\{\vx_t\}_{t \in I_{\mathrm{SR}}}$.
\STATE \textbf{Feature Extraction:}
\STATE $\tilde{\vv}_{k,t}^p = (m^{(k)}_{t-1}, \dots, m^{(k)}_{t-\ell})^\top, \quad \tilde{\vv}_{k,t}^f = (m^{(k)}_{t}, \dots, m^{(k)}_{t+\ell-1})^\top$ for $k=1 \dots m$.
\STATE $\tilde{\phi}^{(k)}(\tilde{\vp}_t) = h(\tilde{\vv}_{k,t}^p), \quad \tilde{\phi}^{(k)}(\tilde{\vf}_t) = h(\tilde{\vv}_{k,t}^f)$ for $k=1 \dots m$.
\STATE Stack into block features: $\tilde{\phi}(\tilde{\vp}_t), \tilde{\phi}(\tilde{\vf}_t) \in \mathbb{R}^m$.
\STATE $\Phi_p = \text{Center}([\tilde{\phi}(\tilde{\vp}_t)]_{t \in I_{\mathrm{SR}}}), \quad \Phi_f = \text{Center}([\tilde{\phi}(\tilde{\vf}_t)]_{t \in I_{\mathrm{SR}}})$.
\STATE \textbf{Subspace Identification:}
\STATE $\mC_- = \frac{1}{|I_{\mathrm{SR}}|} \Phi_p \Phi_p^\top, \quad \mC_+ = \frac{1}{|I_{\mathrm{SR}}|} \Phi_f \Phi_f^\top, \quad \mH = \frac{1}{|I_{\mathrm{SR}}|} \Phi_f \Phi_p^\top$.
\STATE $\mT = \mC_+^{-1/2} \mH \mC_-^{-1/2}$.
\STATE Compute rank-$r$ SVD: $\mT \approx \mU_{l,r} \mathbf{\Sigma}_r \mU_{r,r}^\top$.
\STATE \textbf{State Construction:}
\STATE Canonical direction matrix: $\mathbf{A}_{can} = \mC_-^{-1/2} \mU_{r,r} \in \mathbb{R}^{m \times r}$.
\STATE Past canonical variates: $\vz_{p,t} = \mathbf{A}_{can}^\top \tilde{\phi}(\tilde{\vp}_t)$ for $t \in I_{\mathrm{SR}}$.
\STATE $\vx_t = \mathbf{\Sigma}_r^{1/2} \vz_{p,t}$.
\end{algorithmic}
\end{algorithm}

\subsection{Deep Spectral Encoder Training}
\label{subsec:app-dse-algorithm}

The training process follows a staged schedule to ensure the separation of representation learning and dynamics modeling. Phase 1 employs an alternating optimization with block cross-fitting, while Phase 2 performs end-to-end refinement. The procedure is summarized in Algorithm~\ref{alg:dse-training}. While the theoretical formulation defines the covariance operators without centering, the empirical implementation centers the extracted feature matrices by subtracting their respective sample means before computing the sample covariance blocks and performing the stochastic realization.

\begin{algorithm}[h]
\caption{Training of Deep Spectral Encoder}
\label{alg:dse-training}
\begin{algorithmic}[1]
\REQUIRE Observations $\{\vy_t\}_{t=1}^T$, window $\ell$, state dim $r$, dictionaries $d_A, d_B$, folds $K$.
\ENSURE Parameters $(\eta, \alpha, \theta, \omega, \xi_A, \xi_B)$, operators $\{\hat{\mV}_A, \hat{\mV}_B\}$.
\STATE \textbf{Phase 1: Staged Training with Cross-Fitting}
\STATE Initialize $\eta, \alpha$ (frozen), $\theta, \omega, \xi_A, \xi_B$. Partition $I_{\mathrm{SR}}$ into $\{B_k\}_{k=1}^K$.
\WHILE{not converged}
    \STATE $\{\vx_t\} \leftarrow \text{Stochastic Realization}(\{u_\eta(\vy_t)\}, \ell, r)$ (see Algorithm \ref{alg:sr}).
    \FOR{$k = 1$ \TO $K$}
        \STATE $I_{-k} = I_{\mathrm{SR}} \setminus B_k$.
        \STATE $\hat{\mV}_A^{(-k)} = \Phi^{+}_{-k}(\Phi^{-}_{-k})^{\top}\big(\Phi^{-}_{-k}(\Phi^{-}_{-k})^{\top}+\lambda_A \mI\big)^{-1}$.
        \STATE $\hat{\mV}_B^{(-k)} = \Psi_{-k}(\widehat{\Phi}^{+}_{-k})^{\top}\big(\widehat{\Phi}^{+}_{-k}(\widehat{\Phi}^{+}_{-k})^{\top}+\lambda_B \mI\big)^{-1}$.
        \STATE $\hat{\vh}_{A,t}^{(\mathrm{cf})} = \hat{\mV}_A^{(-k)}\phi_\theta(\vx_{t-1}), \quad \hat{\vh}_{B,t}^{(\mathrm{cf})} = \hat{\mV}_B^{(-k)}\phi_\theta(\hat{\vx}_{t|t-1})$ for $t \in B_k$.
    \ENDFOR
    \STATE Update $(\theta, \omega, \xi_A, \xi_B)$ via Adam on $\mathcal{L}_{\text{rec}}(\hat{\vh}_{A}^{(\mathrm{cf})}, \hat{\vh}_{B}^{(\mathrm{cf})})$.
\ENDWHILE
\STATE \textbf{Phase 2: End-to-End Refinement}
\STATE Unfreeze $\eta, \alpha$.
\WHILE{not converged}
    \STATE Compute $\hat{\vy}_{t|t-1} = g_\alpha(r_{\xi_B}(\hat{\mV}_B \phi_\theta(\hat{\vx}_{t|t-1})))$ for all $t$.
    \STATE Update $(\eta, \alpha, \theta, \omega, \xi_A, \xi_B)$ via Adam on $\sum_t \|\vy_t - \hat{\vy}_{t|t-1}\|^2 - \lambda_{\text{cca}} \sum_{l=1}^r s_l$.
\ENDWHILE
\end{algorithmic}
\end{algorithm}

\subsection{Computational Cost and SVD Implementation}
\label{subsec:app-computational-cost}

We summarize the DSE-specific computational costs of stochastic realization and closed-form operator estimation. Let $N:=|I_{\mathrm{SR}}|$ be the number of valid past--future windows, $\ell$ the window length, $m$ the block-feature dimension, $r$ the retained CCA state dimension, and $d_A,d_B$ the output dimensions of the state and observation dictionaries. Neural feature extraction, block-feature construction, and decoder evaluation are architecture-dependent; Table~\ref{tab:app-complexity} therefore focuses on the computational steps specific to DSE after the relevant features have been evaluated.

\begin{table}[t]
    \caption{DSE-specific computational cost.}
    \label{tab:app-complexity}
    \centering
    \begin{small}
    \begin{tabular}{lll}
        \toprule
        \textbf{Step} & \textbf{Time} & \textbf{Memory} \\
        \midrule
        CCA covariance blocks $\mC_-,\mC_+,\mH$ 
        & $\mathcal{O}(Nm^2)$ 
        & $\mathcal{O}(m^2)$ \\
        CCA whitening and SVD 
        & $\mathcal{O}(m^3)$ 
        & $\mathcal{O}(m^2)$ \\
        CCA projection 
        & $\mathcal{O}(Nmr)$ 
        & $\mathcal{O}(Nr)$ \\
        Transfer solve $\hat{\mV}_A$ 
        & $\mathcal{O}(Nd_A^2+d_A^3)$ 
        & $\mathcal{O}(d_A^2)$ \\
        Observation solve $\hat{\mV}_B$ 
        & $\mathcal{O}(N(d_A^2+d_A d_B)+d_A^3+d_A^2d_B)$ 
        & $\mathcal{O}(d_A^2+d_A d_B)$ \\
        \bottomrule
    \end{tabular}
    \end{small}
\end{table}

The main bottlenecks are the covariance formation and the CCA whitening/SVD when the block-feature dimension $m$ is large. The window length $\ell$ affects the architecture-dependent block-feature construction cost and enters the computational cost through the chosen block-feature dimension $m$. The operator-estimation steps work with feature- and dictionary-dimensional matrices and do not construct an $N\times N$ sample-indexed Gram matrix. If intermediate features are cached, the additional storage is linear in $N$.

The SVD in Algorithm~\ref{alg:sr} solves the finite-dimensional CCA problem for the normalized past--future cross-covariance matrix. The retained rank-$r$ singular directions define the predictive state coordinates used by stochastic realization. During representation learning, this CCA/SVD block is differentiable: gradients can be propagated through the rank-$r$ SVD to the normalized cross-covariance matrix and then to the block-feature and encoder parameters. This differentiation is used for learning the representation; the transfer and observation operators are still obtained by the ridge-regression updates in Eq.~\eqref{eq:cf-va} and Eq.~\eqref{eq:cf-vb} rather than treated as unconstrained recurrent parameters. Numerically, the whitening step is regularized by diagonal loading, and only a fixed rank-$r$ subspace is retained, which truncates directions associated with small or unstable singular values.

\subsection{Sequential State Estimation (Filtering)}
\label{subsec:app-filtering-algorithm}

While the underlying dynamical system is nonlinear and stochastic, DSE enables sequential inference via linear recursions within the learned Hilbert space embeddings. Algorithm~\ref{alg:filtering} summarizes the noise statistics estimation and the recursive filtering process.

\paragraph{Empirical Noise Estimation.}
The filtering process requires the process and observation noise covariances, $\mathcal{Q}$ and $\mathcal{R}$. In our framework, these are estimated empirically from the residuals of the learned operators on the training data. This approach accounts for the approximation error of the Galerkin projection as a form of ``modeling noise.'' The state residuals $\bm e^x_t$ and observation residuals $\bm e^y_t$ are computed from the difference between the one-step operator predictions and the actual encoded features.

\paragraph{Feature-Space Recursion.}
The recursion implemented in Algorithm~\ref{alg:filtering} propagates the first and second moments of the embedded state distribution. Unlike standard Kalman filters that operate on the observation space, this filter acts on the feature coordinates $\bm{\phi}_t$. The ``Innovation'' step incorporates new information by projecting the observation into the feature space via $\psi_\omega(u_\eta(\vy_t))$, allowing for nonlinear updates to the latent state while maintaining a linear computational structure.

\begin{algorithm}[h]
\caption{Feature-Space Sequential State Estimation}
\label{alg:filtering}
\begin{algorithmic}[1]
\REQUIRE Learned DSE, new observations $\{\vy_t\}_{t=1}^T$, regularization $\gamma_{\mathcal{Q}}, \gamma_{\mathcal{R}}$.
\ENSURE State estimates $\{\hat{\vx}_t\}_{t=1}^T$.
\STATE \textbf{Noise Estimation:}
\STATE $\bm e^x_t = \phi_\theta(\vx_t) - \hat{\mV}_A \phi_\theta(\vx_{t-1}), \quad
\bm e^y_t = \psi_\omega(\vm_t) - \hat{\mV}_B \phi_\theta(\vx_t)$.
\STATE $\mathcal{Q} = \frac{1}{|I_{\mathrm{SR}}|-1} \sum_t \bm e^x_t(\bm e^x_t)^\top + \gamma_{\mathcal{Q}} \mI, \quad
\mathcal{R} = \frac{1}{|I_{\mathrm{SR}}|} \sum_t \bm e^y_t(\bm e^y_t)^\top + \gamma_{\mathcal{R}} \mI$.
\STATE \textbf{Recursion:} 
\STATE Initialize $\bm{\mu}_0^+, \mathbf{\Sigma}_0^+$ from training ensemble.
\FOR{$t = 1$ \TO $T$}
    \STATE $\bm{\mu}_t^- = \hat{\mV}_A \bm{\mu}_{t-1}^+, \quad \mathbf{\Sigma}_t^- = \hat{\mV}_A \mathbf{\Sigma}_{t-1}^+ \hat{\mV}_A^\top + \mathcal{Q}$.
    \STATE $\hat{\bm{\psi}}_t^- = \hat{\mV}_B \bm{\mu}_t^-, \quad \mathbf{S}_t = \hat{\mV}_B \mathbf{\Sigma}_t^- \hat{\mV}_B^\top + \mathcal{R}$.
    \STATE $\mK_t = \mathbf{\Sigma}_t^- \hat{\mV}_B^\top \mathbf{S}_t^{-1}$.
    \STATE $\bm{\psi}_t = \psi_\omega(u_\eta(\vy_t))$.
    \STATE $\bm{\mu}_t^+ = \bm{\mu}_t^- + \mK_t (\bm{\psi}_t - \hat{\bm{\psi}}_t^-)$.
    \STATE $\mathbf{\Sigma}_t^+ = (\mI - \mK_t \hat{\mV}_B) \mathbf{\Sigma}_t^-$.
    \STATE $\hat{\vx}_t = r_{\xi_A}(\bm{\mu}_t^+)$.
\ENDFOR
\end{algorithmic}
\end{algorithm}

\section{Training Details}

\subsection{Cross Fitting}

To prevent self‑fitting and temporal leakage when estimating the finite-dimensional transfer operators and learning the read-out maps, we adopt block cross‑fitting over contiguous time segments. We partition the time index set $\{1,\, \dots, \, T\}$ into $K$ disjoint, contiguous blocks $\{B_k\}_{k=1}^K$ and define the out‑of‑fold time index set for block $B_k$ by $I_{-k}:= \{1,\, \dots, \, T\} \backslash B_k$. In all experiments, we use $K=5$ contiguous blocks with a minimum block size of $100$ time steps. All stage‑1 estimators are fitted on $I_{-k}$, and the corresponding out‑of‑fold features are then computed on the held‑out block $B_k$. This construction preserves the test‑time causal path and removes information flow from $B_k$ into its own predictors. The same scheme is applied to the estimation of both the state process transfer operator  and the observation operator.

Let $\phi_\theta : \R^r\rightarrow \R^{d_A}$ be the state feature map, $\mV_A : \R^{d_A}\rightarrow \R^{d_A}$ the transfer operator in the feature space, and $r_{\xi_A} : \R^{d_A}\to \R^r$ the state read-out map. For each fold $k$, stage-1 fits an out‑of‑fold operator $\mV_A^{(-k)}$ by ridge-regularized multi-target least squares problem on $I_{-k}$:
\begin{align}
    \mV_A^{(-k)} = \Phi^{+}_{-k}(\Phi^{-}_{-k})^{\top}\big(\Phi^{-}_{-k}(\Phi^{-}_{-k})^{\top}+\lambda_A \mI\big)^{-1}, 
\end{align}
where $\Phi_{-k}^- = [\phi_\theta(\vx_{t-1})]_{t\in I_{-k}} \in \R^{d_A\times |I_{-k}|}$ and $\Phi_{-k}^+ = [\phi_\theta(\vx_{t})]_{t\in I_{-k}} \in \R^{d_A\times |I_{-k}|}$. Out-of-fold features on $B_k$ are then computed as
\begin{align}
    \widehat \mH^{({\mathrm{cf}})}_{A,B_k} := \mV_A^{(-k)}\,\Phi^{-}_{B_k}
\end{align}
where $\Phi^{-}_{B_k} = \big[\phi_\theta(\vx_{t-1})\big]_{t\in B_k}$. Let $\widehat \mH_A^{(\mathrm{cf})}\in \R^{d_A\times T}$ denote the out-of-fold design matrix obtained by stacking the block-wise matrices $\widehat \mH_{A, B_k}^{(\mathrm{cf})}$ column-wise over all folds, such as
\begin{align}
    \widehat \mH_A^{(\mathrm{cf})} = [\widehat \mH_{A, B_1}^{(\mathrm{cf})},\, \dots,\, \widehat \mH_{A, B_K}^{(\mathrm{cf})} ].
\end{align}
Then stage-2 learns the state read-out map parameters $\xi_A$ only from the out-of-fold design:
\begin{align}
\widehat \xi_A
=\underset{\xi_A}{\mathrm{argmin}}\ \frac{1}{\mid I_\mathrm{SR} \mid}\sum_{t\in I_\mathrm{SR}}
\left\|\vx_t-r_{\xi_A}\!\left(\widehat \vh^{(\mathrm{cf})}_{A,t}\right)\right\|_2^2
+\lambda_{r,A}\Omega(\xi_A),
\end{align}
where $\widehat \vh^{(\mathrm{cf})}_{A,t}$ denotes the $t$-th column of $\widehat \mH^{(\mathrm{cf})}_A$; $\Omega(\cdot)$ denotes a regularizer on the read-out parameters and $\lambda_{r,A}$ is the corresponding regularization coefficient. This cross‑fitted two‑stage estimator preserves the computation graph at the inference one-step prediction $\hat \vx_{t\,|\,t-1}$ and mitigates co-adaptation between operators and neural networks as encoder/decoder.

For the observation operator we use the same block cross-fitting protocol as for the state operator. Let $\psi_\omega:\mathbb{R}^{m}\to\mathbb{R}^{d_B}$ be the observation feature map, $\mV_B:\mathbb{R}^{d_A}\to\mathbb{R}^{d_B}$ the observation operator in feature space, and $r_{\xi_B}:\mathbb{R}^{d_B}\to\mathbb{R}^{m}$ the observation read-out map. For each fold $k$ we consider the out-of-fold index set $I_{-k}$ and the corresponding held-out block $B_k$ defined above. On $I_{-k}$ we form the feature matrix for state process as $\Phi_{-k}=[\phi_\theta(\hat\vx_{t|t-1})]_{t\in I_{-k}}\in \R^{d_A \times |I_{-k}|}$ and the feature matrix for observations as $\Psi_{-k}=[\psi_\omega(\vm_t)]_{t\in I_{-k}}\in \R^{d_B \times |I_{-k}|}$. The first stage then estimates an out-of-fold matrix representation for the observation operator $\mV_B^{(-k)}$ by ridge-regularized least squares,
\begin{align}
    \mV_B^{(-k)}= \Psi_{-k}\,\Phi_{-k}^\top\bigl(\Phi_{-k}\Phi_{-k}^\top + \lambda_B \mI\bigr)^{-1},
\end{align}
which has the same closed form as in the main paper but is fitted only on indices in $I_{-k}$. On the held-out block $B_k$ we construct the corresponding out-of-fold design by propagating the one-step state predictors through $\mV_B^{(-k)}$, that is,
\begin{align}
\Phi_{B_k}= \big[\phi_\theta(\hat \vx_{t\mid t-1})\big]_{t\in B_k},
\qquad
\widehat \mH^{(\mathrm{cf})}_{B,B_k}=\mV_B^{(-k)}\,\Phi_{B_k}.
\end{align}
Arranging the block-wise matrices $\widehat \mH^{(\mathrm{cf})}_{B,B_k}$ column-wise over all folds yields the global out-of-fold design $\widehat \mH^{(\mathrm{cf})}_{B}\in \R^{d_B \times T}$ used in Stage-2. Let $\mM = [\vm_t]_{t\in I_\mathrm{SR}}$ denote the encoder feature matrix whose columns are the encoder features $\vm_t$. The observation read-out map parameters $\xi_B$ are learned only from the out-of-fold design:
\begin{align}
\widehat \xi_B
=\underset{\xi_B}{\mathrm{argmin}}\ \frac{1}{|I_\mathrm{SR}|}\sum_{t\in I_\mathrm{SR}}
\left\|\vm_t-r_{\xi_B}\!\left(\widehat \vh^{(\mathrm{cf})}_{B,t}\right)\right\|_2^2
+\lambda_{r,B}\Omega(\xi_B),
\end{align}
where $\widehat \vh^{(\mathrm{cf})}_{B,t}$ denotes the $t$-th column of $\widehat \mH^{(\mathrm{cf})}_B$ and $\lambda_{r,B}$ is the corresponding regularization coefficient. By this construction, each column of $\widehat \mH^{(\mathrm{cf})}_B$ is generated by an operator $\mV_B^{(-k)}$ that has never been fitted on its own time index, so the features entering the second-stage regression are genuinely out-of-fold.

This separation helps to mitigate co-adaptation between the encoder and the observation operator and encourages $\mV_B$ to act as a stable, time-invariant linear map from the state feature subspace induced by $\phi_\theta$ into the observation feature subspace induced by $\psi_\omega$, rather than drifting toward a nearly trivial transfer matrix.



\subsection{Ablation Study}
\label{subsec:app-ablation}

We report two ablations on the quad-link pendulum image benchmark used in Section~\ref{sec:application:seq}. The first ablation compares the staged optimization schedule with joint end-to-end training, and the second isolates the contribution of the main DSE components on the $1.5\times 10^3$ corrupted regime. The data generation, encoder--decoder architectures, and regularization coefficients are kept identical to the main experiment unless otherwise stated.

\paragraph{Training Schedule.}


In the staged schedule, operator estimation is decoupled from representation learning. We first fix the encoder $u_\eta$ and decoder $g_\alpha$, estimate the operators $\mV_A$ and $\mV_B$ by solving regularized least-square problems corresponding to the Galerkin projections on the Hilbert spaces spanned by the feature maps $\phi_\theta$ and $\psi_\omega$, and learn the read-out maps $r_{\xi_A}$ and $r_{\xi_B}$ by minimizing the corresponding regularized objectives on the operator-projected features. After updating the operators and the read-out maps in the first phase, we further optimize the neural components along the same training objective as in the main experiment, namely the reconstruction loss augmented by a regularization term given by the sum of canonical correlation coefficients obtained from the operator-based stochastic realization. In this refinement phase, $\mV_A$ and $\mV_B$ may be kept fixed or recomputed from the current features, and the read-out maps may either be fixed or fine-tuned depending on the selected implementation. Under this schedule the temporal structure is primarily encoded through $\mV_A$ and $\mV_B$, whereas the encoder and decoder are encouraged to produce feature coordinates and reconstructions that are compatible with the latent Markov process induced by the staged operator estimation rather than directly absorbing sequence-level dynamics.

In the joint schedule, all modules $(\eta,\theta,\omega,\alpha,\xi_A,\xi_B)$ and the operators are trained simultaneously from scratch with the same reconstruction loss augmented with a canonical-correlation regularizer. At each gradient step, the closed-form solutions for $\mV_A$ and $\mV_B$ are recomputed from the current features and treated as differentiable functions of $\phi_\theta$ and $\psi_\omega$. This tight coupling tends to increase gradient interference between representation learning and operator estimation, so that the encoder absorbs much of the temporal structure into its own representation.


\begin{table}[t]
    \caption{Ablation Study on Training Schedules Using Quad-Link Pendulum Dataset}
    \label{tab:ablation_schedule}
    \centering
    \begin{small}
    \begin{tabular}{lcccc}
        \toprule
        & \multicolumn{2}{c}{\textbf{Length 1.5k}} & \multicolumn{2}{c}{\textbf{Length 15k}} \\
        \cmidrule(lr){2-3} \cmidrule(lr){4-5}
        \textbf{Schedule} & \textbf{No noise} & \textbf{With noise} & \textbf{No noise} & \textbf{With noise} \\
        \midrule
        \textbf{Staged (Proposed)} & $\bm{0.2006} \pm 0.0228$ & $\bm{0.2593} \pm 0.0274$ & $\bm{0.2615} \pm 0.0211$ & $\bm{0.2683} \pm 0.0254$ \\
        Joint & $0.2248 \pm 0.0378$ & $0.2977 \pm 0.0512$ & $0.2835 \pm 0.0344$ & $0.3202 \pm 0.0474$ \\
        \bottomrule
    \end{tabular}
    \end{small}
\end{table}

Table~\ref{tab:ablation_schedule} reports the one-step prediction MSE of DSE under the staged and joint schedules on quad-link pendulum images, for training sequence lengths $1.5\times 10^3$ and $1.5\times 10^4$ and for both the clean and corrupted regimes. Across these settings, the staged schedule attains lower or comparable error, with the largest gap typically observed for long corrupted sequences where joint training frequently converges to degenerate operator solutions. These results support the use of the staged schedule as the default optimization strategy in the main experiments.

\paragraph{Component Ablation.}
We further isolate the contribution of the main DSE components on the $1.5\times 10^3$ corrupted regime. The variant without CCA-based state construction uses the encoder features directly as state coordinates, bypassing the stochastic realization step. The variant without closed-form operator estimation replaces the ridge-regression solutions for $\mV_A$ and $\mV_B$ with three-layer neural networks parameterizing the transition and observation maps, trained by gradient-based optimization. The joint-training variant is the same as the joint schedule above and removes the staged optimization procedure. The role of learned deep features is assessed by the ELTO-KF comparison in Table~\ref{tab:quadlink}, which follows the same broad operator-based filtering framework but does not learn the explicit neural feature and dictionary spaces used by DSE.

\begin{table}[t]
    \caption{Ablation Study on Components Using the $1.5\times 10^3$ Corrupted Quad-Link Benchmark}
    \label{tab:ablation_components}
    \centering
    \begin{small}
    \begin{tabular}{lcc}
        \toprule
        \textbf{Variant} & \textbf{One-step MSE} & \textbf{Multi-step MSE at $H=10$} \\
        \midrule
        \textbf{Full DSE} & \textbf{0.2593} $\pm$ 0.0274 & \textbf{0.2952} $\pm$ 0.0357 \\
        w/o CCA-based state construction & 0.2903 $\pm$ 0.0455 & 0.3406 $\pm$ 0.0551 \\
        w/o closed-form operator estimation & 0.2758 $\pm$ 0.0405 & 0.3203 $\pm$ 0.0505 \\
        joint training instead of staged training & 0.2977 $\pm$ 0.0512 & 0.3553 $\pm$ 0.0752 \\
        \bottomrule
    \end{tabular}
    \end{small}
\end{table}

Table~\ref{tab:ablation_components} reports the one-step prediction MSE and the multi-step prediction MSE at horizon $H=10$ for these variants. Removing any of the main components degrades prediction relative to the full DSE. The degradation under the variant without CCA-based state construction is more pronounced for multi-step prediction, indicating that the predictive state coordinates obtained by stochastic realization contribute to rollout stability. Replacing the closed-form operator estimation with neural-network parameterization increases the variability across runs, while joint training gives the largest degradation among the tested variants.

\section{Experimental Setting}

This section specifies the data generation procedures, noise settings, baseline configurations, and evaluation protocols used in the experiments.

\subsection{Sequential State Estimation}
\label{subsec:aapdx-sse}

We evaluate sequential prediction on simulated quad-link pendulum trajectories observed as $48 \times 48$ grayscale image sequences. Initial joint angles are sampled uniformly from $[-\pi,\,\pi]$ with zero initial velocities, the viscous friction coefficient is set to $0.1$, the simulator time step is $10^{-4}$ seconds, and frames are recorded every $0.05$ seconds. In the corrupted regime, we use the fixed noise configuration with $\rho=0.2$ and thresholds $[0,\,0.25,\,0.75,\,1.0]$, and we keep the first five frames in each sequence clean. We report prediction mean squared error for the two sequence-length settings $1.5\times10^3$ and $1.5\times10^4$, and Table~\ref{tab:quadlink} summarizes the mean and standard deviation over $5$ independent runs.

All methods use the same image resolution, preprocessing, simulator configuration, and data split. The quad-link sequences are generated using the simulator distributed in the RKN reference repository at \url{https://github.com/ALRhub/rkn-share}, and the remaining simulator parameters follow the simulator defaults.

\paragraph{Simulation and Corruption Setting.}
Table~\ref{tab:quadlink_setting} summarizes the quad-link data generation, corruption process, and evaluation protocol.
We follow the simulator configuration provided in the RKN reference implementation \citep{BPG+19} and adopt the same data
split protocol as in the setting of ELTO-based method \citep{nry2025elto}. Specifically, we use $1,200$ trajectories to train
our method, $300$ trajectories to train the decoder, and $200$ trajectories for validation and evaluation.

\begin{table}[t]
  \centering
  \small
  \setlength{\tabcolsep}{4pt}
  \caption{Quad-Link Data Generation and Corruption Setting.}
  \label{tab:quadlink_setting}
  \begin{tabular}{@{}p{0.44\linewidth}p{0.52\linewidth}@{}}
    \hline
    \textbf{Item} & \textbf{Setting and Values} \\
    \hline
    \multicolumn{2}{@{}l}{\textbf{Simulation / observation}} \\
    Image resolution & $48 \times 48$ grayscale \\
    Initial joint angles & $q(0)\sim \mathrm{Unif}([-\pi,\pi])$;\; $\dot q(0)=0$ \\
    Friction coefficient & $0.1$ \\
    Simulator time step & $dt=10^{-4}\,\mathrm{s}$ \\
    Frame interval & $\Delta t_{\mathrm{frame}}=0.05\,\mathrm{s}$ \\
    \hline
    \multicolumn{2}{@{}l}{\textbf{Sequence / split}} \\
    Sequence lengths evaluated & $1.5\times 10^{3}$ and $1.5\times 10^{4}$ frames \\
    \# trajectories (train / val ) & $1200 / 200 $ \\
    \# trajectories for decoder (if applicable) & $300$ \\
    \hline
    \multicolumn{2}{@{}l}{\textbf{Corruption setting}} \\
    Correlation parameter & $\rho=0.2$ \\
    Thresholds & $[0,\;0.25,\;0.75,\;1.0]$ \\
    Clean prefix & First $5$ frames are kept clean \\
    \hline
    \multicolumn{2}{@{}l}{\textbf{Evaluation}} \\
    Metric & Prediction mean squared error \\
    Repeats & $5$ independent runs (mean $\pm$ std.) \\
    \hline
  \end{tabular}
\end{table}

\paragraph{Parameters for Our Method.}
Table~\ref{tab:quadlink_params} reports the architecture and training hyperparameters of our method used in the quad-link
experiment. We use the staged training schedule described in the main text. To ensure numerical stability when computing the normalized cross-covariance $\tilde\mT = (\mC_+ + \delta_{\mathrm{cca}} \mI)^{-1/2} \,\mH\, (\mC_- + \delta_{\mathrm{cca}} \mI)^{-1/2}$, we apply diagonal loading with $\delta_{\mathrm{cca}} > 0$ as the covariance regularization. Furthermore, in our experiments, we set $\gamma_{\mathcal Q}=\gamma_{\mathcal R}=10^{-6}$ in the clean regime and
$\gamma_{\mathcal Q}=\gamma_{\mathcal R}=10^{-3}$ in the corrupted regime.

\begin{table}[t]
  \centering
  \small
  \setlength{\tabcolsep}{4pt}
  \caption{Hyperparameters for Our Method in Quad-Link Experiment.}
  \label{tab:quadlink_params}
  \begin{tabular}{@{}p{0.44\linewidth}p{0.52\linewidth}@{}}
    \hline
    \textbf{Component / hyperparameter} & \textbf{Architectures and Values} \\
    \hline
    \multicolumn{2}{@{}l}{\textbf{Encoder $u_{\eta}$ }} \\
    Convolution 1 &
      12 filters, $5\times 5$ kernel, stride $2\times 2$, ReLU;
      $2\times 2$ max-pooling with stride $2\times 2$ \\
    Convolution 2 &
      12 filters, $3\times 3$ kernel, stride $2\times 2$, ReLU;
      $2\times 2$ max-pooling with stride $2\times 2$ \\
    Fully connected & 200 units, ReLU \\
    Output heads &
      Mean head: linear; variance head: $\mathrm{ELU}(\cdot)+1$ to enforce nonnegative variance \\
    Padding / normalization &
      Zero-padding is chosen so that each convolution preserves the spatial resolution;
      layer normalization is applied in all convolutional layers \\
    Feature dimension $m$ & $ 100$ \\
    \hline
    \multicolumn{2}{@{}l}{\textbf{Decoder $g_{\alpha}$ }} \\
    Decoder architecture &
      Fully connected: 144, ReLU;
      transposed conv: 16 filters, $5\times 5$ kernel, stride $4\times 4$, ReLU;
      transposed conv: 12 filters, $3\times 3$ kernel, stride $4\times 4$, ReLU;
      transposed conv output: 1 channel, stride $1\times 1$, Sigmoid \\
    Output type & Sigmoid output (image intensity in $(0,1)$) \\
    \hline
    \multicolumn{2}{@{}l}{\textbf{Deep spectral learning / state model}} \\
    Latent state dimension $r$ & 20\\
    SR window length $\ell$ & 30\\
    Dictionary dimensions $d_A, d_B$ & $50\, / \,50$\\
    State dictionary $\phi_\theta$ (DF-A feature net) &
      MLP: $20 \rightarrow 128 \rightarrow 64 \rightarrow 50$, ReLU, dropout $0.1$\\
    Observation dictionary $\psi_\omega$ (DF-B obs net) &
      MLP: $100 \rightarrow 64 \rightarrow 32 \rightarrow 50$, ReLU, dropout $0.1$\\
    Feature mapping MLP $h$ (encoder output $\rightarrow$ block feature) &
      1-hidden-layer MLP: $\ell(=30)\rightarrow 32 \rightarrow 1$, ReLU\\
    Read-out map $r_{\xi_A}$ &
      2-hidden-layer MLP: $50 \rightarrow 32 \rightarrow 32 \rightarrow 20$, ReLU\\
    Read-out map $r_{\xi_B}$ &
      2-hidden-layer MLP: $50 \rightarrow 32 \rightarrow 32 \rightarrow 100$, ReLU\\
    \hline
    \multicolumn{2}{@{}l}{\textbf{Regularization}} \\
    Ridge regularization $\lambda_A, \lambda_B$ & $10^{-2}\,/\,10^{-3}$\\
    Covariance regularization $\delta_{\mathrm{cca}}$ (diagonal loading) & $10^{-3}$\\
    CCA regularization weight $\lambda_{\mathrm{cca}}$ & $10^{-4}$\\
    \hline
    \multicolumn{2}{@{}l}{\textbf{Optimization (staged schedule)}} \\
    Optimizer & Adam\\
    Learning rate ($u_\eta, g_\alpha, \phi_\theta, \psi_\omega$) & $10^{-3}$\\
    Batch size & 150\\
    Stage-1 epochs & 25\\
    Stage-2 epochs & 100\\
    Cross-fitting (contiguous blocks) & $K=5$, min\_block\_size $=100$\\
    \hline
    \multicolumn{2}{@{}l}{\textbf{Compute / reproducibility}} \\
    Framework / hardware & Tesla V100S-PCIE-32GB GPU \\
    \hline
  \end{tabular}
\end{table}

\subsection{Mode Decomposition}
\label{subsec:app-md}

We evaluate Koopman spectrum recovery on the Van der Pol (VDP) oscillator with additive observation noise and the Stuart-Landau (SL) oscillator with additive process noise, defined in Eq.~\eqref{eq:vdp} and~\eqref{eq:sl}. For each noise level, we compute the absolute eigenvalue error between the estimated spectrum and a reference spectrum, and report mean $\pm$ standard deviation over 50 independent trials. To ensure reproducibility, we summarize the detailed simulation configurations and the hyperparameters of the proposed method in the following tables. For the oscillator experiments, although the simulator outputs scalar observations, we use time-delay embedding to construct vector-valued inputs to the encoder and decoder.


\subsubsection{Van der Pol Oscillator with Observation Noise}
\label{subsec:app-vdp}


\begin{table}[!t]
  \centering
  \small
  \setlength{\tabcolsep}{4pt}
  \caption{Simulation Setting for VDP Oscillator Experiment}
  \label{tab:vdp_setting}
  \begin{tabular}{@{}p{0.44\linewidth}p{0.52\linewidth}@{}}
    \hline
    \textbf{Item} & \textbf{Setting and Values} \\
    \hline
    Parameter & $\mu = 2.0$ \\
    Integrator / step size & 4th-order Runge--Kutta, $\Delta t = 0.1$ \\
    Initial condition & $(x_0, y_0) = (2.0, 0.0)$ \\
    Trajectory length & $T = 3{,}500$ \\
    Train/validation split & first $3{,}000$ / last $500$ samples \\
    Observation & scalar observation with additive noise \\
    Observation noise & i.i.d.\ Gaussian, mean $0$, variance $\sigma_{\mathrm{obs}}^{2}$ \\
    Noise sweep & $\sigma_{\mathrm{obs}}^{2} \in \{0.01, 0.02, \ldots, 0.20\}$ \\
    Trials per noise level & 50 independent trajectories \\
    Reference spectrum & integer harmonics of the frequency (noise-free); $\omega = 0.823498$ \\
    Metric / reporting & absolute eigenvalue error; mean $\pm$ std over $50$ trials \\
    \hline
  \end{tabular}
\end{table}

\begin{table}[!t]
  \centering
  \small
  \setlength{\tabcolsep}{4pt}
  \caption{Parameter Setting for Our Method in VDP Oscillator Experiment.}
  \label{tab:vdp_params}
  \begin{tabularx}{1.0\linewidth}{@{}lX@{}}
    \hline
    \textbf{Component / hyperparameter} & \textbf{Architectures and values} \\
    \hline
    \multicolumn{2}{@{}l}{\textbf{Encoder $u_{\eta}$}} \\
    Encoder architecture &
      4-layer MLP with $\tanh$: $30 \rightarrow 128 \rightarrow 128 \rightarrow 64 \rightarrow 50$ \\
    Feature dimension $m$ & 50 \\
    \hline
    \multicolumn{2}{@{}l}{\textbf{Decoder $g_{\alpha}$}} \\
    Decoder architecture &
      4-layer MLP with $\tanh$: $50 \rightarrow 64 \rightarrow 128 \rightarrow 128 \rightarrow 30$ (linear output) \\
    \hline
    \multicolumn{2}{@{}l}{\textbf{Deep spectral learning / state model}} \\
    Latent state dimension $r$ & 9 \\
    SR window length $\ell$ & 20 \\
    Dictionary dimensions $d_A, d_B$ & 30 / 30 \\
    State dictionary $\phi_\theta$ (DF-A feature net) &
      MLP: $9 \rightarrow 128 \rightarrow 64 \rightarrow 30$, ReLU, dropout $0.1$\\
    Observation dictionary $\psi_\omega$ (DF-B obs net) &
      MLP: $50 \rightarrow 64 \rightarrow 32 \rightarrow 30$, ReLU, dropout $0.1$\\
    Feature mapping MLP $h$ (encoder output $\rightarrow$ block feature) &
      1-hidden-layer MLP: $\ell(=20)\rightarrow 32 \rightarrow 1$, ReLU\\
    Read-out map $r_{\xi_A}$ &
      2-hidden-layer MLP: $30 \rightarrow 32 \rightarrow 32 \rightarrow 9$, ReLU\\
    Read-out map $r_{\xi_B}$ &
      2-hidden-layer MLP: $30 \rightarrow 32 \rightarrow 32 \rightarrow 50$, ReLU\\
    \hline
    \multicolumn{2}{@{}l}{\textbf{Regularization}} \\
    Ridge regularization $\lambda_A, \lambda_B$ & $10^{-2} / 10^{-3}$ \\
    Covariance regularization $\delta_{\mathrm{cca}}$ (diagonal loading) & $10^{-3}$\\
    CCA regularization weight $\lambda_{\mathrm{cca}}$ & $10^{-3}$\\
    \hline
    \multicolumn{2}{@{}l}{\textbf{Optimization (staged schedule)}} \\
    Optimizer & Adam \\
    Learning rate ($u_\eta, g_\alpha$) & $10^{-4}$\\
    Learning rate ($\phi_\theta, \psi_\omega$) & $10^{-3}$\\
    Batch size & 100 \\
    Stage-1 epochs & 25 \\
    Stage-2 epochs & 100 \\
    Cross-fitting (contiguous blocks) & $K=5$, min\_block\_size $=100$\\
    \hline
    \multicolumn{2}{@{}l}{\textbf{Compute / reproducibility}} \\
    Framework / hardware & Tesla V100S-PCIE-32GB GPU \\
    \hline
  \end{tabularx}
\end{table}

\paragraph{Governing Equations and Simulation.}
This section details the data generation process for the VDP oscillator evaluated in Section~\ref{sec:application:mode}. The deterministic continuous-time system is governed by
\begin{align}
    \frac{dx}{dt} = y,\quad \frac{dy}{dt} = \mu (1-x^2)y - x,
    \label{eq:vdp}
\end{align}
where the nonlinearity parameter $\mu$ is defined in Table~\ref{tab:vdp_setting}. Following the simulation configuration used in the ELTO supplementary setting \citep{nry2025elto}, we integrate this system numerically using the fourth-order Runge-Kutta method. The precise simulation parameters, including the initial conditions, time step $\Delta t$, trajectory lengths, and training splits, are summarized in Table~\ref{tab:vdp_setting}.

\paragraph{Noise Setting and Evaluation Protocol.}
To rigorously evaluate robustness, we corrupt the observations with independent and identically distributed zero-mean Gaussian noise having variance $\sigma^2_\mathrm{obs}$. As outlined in Section~\ref{sec:application:mode}, we sweep $\sigma^2_\mathrm{obs}$ from $0.01$ to $0.20$ with an increment of $0.01$. The number of independent trials generated per noise level is provided in Table~\ref{tab:vdp_setting}. For the evaluation protocol, the reference spectrum is constructed using the integer harmonics of the noise-free limit-cycle frequency $\omega = 0.823498$, which corresponds to extracting the discrete-time eigenvalues at the sampling interval $\Delta t$.

\paragraph{Parameters for our method.}
Table~\ref{tab:vdp_params} reports the architecture and training hyperparameters of our method used in the VDP experiment. As introduced in Appendix~\ref{subsec:aapdx-sse}, we use the same diagonal loading $\delta_{\mathrm{cca}}> 0$ for the covariance regularization to ensure numerical stability.

\begin{table}[!t]
  \centering
  \small
  \setlength{\tabcolsep}{4pt}
  \caption{Simulation Setting for SL Oscillator Experiment}
  \label{tab:sl_setting}
  \begin{tabular}{@{}p{0.44\linewidth}p{0.52\linewidth}@{}}
    \hline
    \textbf{Item} & \textbf{Value} \\
    \hline
    Parameters & $\mu = 1.0,\ \gamma = 0.9,\ \beta = 0.3$ \\
    Integrator / step size & 4th-order Runge--Kutta, $\Delta t = 0.1$ \\
    Initial condition & $(r_0, \theta_0) = (0.1, 0.0)$ \\
    Trajectory length & $T = 3{,}500$ \\
    Train/validation split & first $3{,}000$ / last $500$ samples \\
    Observation & $y_t = e^{i\theta_t}$ \\
    Observation noise & i.i.d.\ Gaussian, variance fixed to $0.01$ \\
    Process noise sweep & variance $\epsilon_{\mathrm{proc}} \in \{0.01, 0.02, \ldots, 0.09\}$ \\
    Trials per noise level & $50$ independent trajectories \\
    Reference spectrum & asymptotic theory under weak process noise \citep{bag2014};\\
    &$\omega=0.6$, $\kappa=3.0$ \\
    Metric / reporting & absolute eigenvalue error; mean $\pm$ std over 50 trials \\
    \hline
  \end{tabular}
\end{table}

\begin{table}[!t]
  \centering
  \small
  \setlength{\tabcolsep}{4pt}
  \caption{Parameter Setting for Our Method in SL Oscillator Experiment.}
  \label{tab:sl_params}
  \begin{tabularx}{1.0\linewidth}{@{}lX@{}}
    \hline
    \textbf{Component / hyperparameter} & \textbf{Architectures and values} \\
    \hline
    \multicolumn{2}{@{}l}{\textbf{Encoder $u_{\eta}$}} \\
    Encoder architecture &
      4-layer MLP with $\tanh$: $20 \rightarrow 128 \rightarrow 128 \rightarrow 64 \rightarrow 50$ \\
    Feature dimension $m$ & 50 \\
    \hline
    \multicolumn{2}{@{}l}{\textbf{Decoder $g_{\alpha}$}} \\
    Decoder architecture &
      4-layer MLP with $\tanh$: $50 \rightarrow 64 \rightarrow 128 \rightarrow 128 \rightarrow 20$ (linear output) \\
    \hline
    \multicolumn{2}{@{}l}{\textbf{Deep spectral learning / state model}} \\
    Latent state dimension $r$ & 13 \\
    SR window length $\ell$ & 20 \\
    Dictionary dimensions $d_A, d_B$ & 30 / 30 \\
    State dictionary $\phi_\theta$ (DF-A feature net) &
      MLP: $13 \rightarrow 128 \rightarrow 64 \rightarrow 30$, ReLU, dropout $0.1$\\
    Observation dictionary $\psi_\omega$ (DF-B obs net) &
      MLP: $50 \rightarrow 64 \rightarrow 32 \rightarrow 30$, ReLU, dropout $0.1$\\
    Feature mapping MLP $h$ (encoder output $\rightarrow$ block feature) &
      1-hidden-layer MLP: $\ell(=20)\rightarrow 64 \rightarrow 1$, ReLU\\
    Read-out map $r_{\xi_A}$ &
      2-hidden-layer MLP: $30 \rightarrow 32 \rightarrow 32 \rightarrow 13$, ReLU\\
    Read-out map $r_{\xi_B}$ &
      2-hidden-layer MLP: $30 \rightarrow 32 \rightarrow 32 \rightarrow 50$, ReLU\\
    \hline
  \end{tabularx}
\end{table}

\subsubsection{Stuart-Landau Oscillator with Process Noise}
\label{subsec:app-sl}

\paragraph{Governing Equations and Observation.}
This section details the explicit complex-valued differential equations and observation functions for SL oscillator evaluated in Section~\ref{sec:application:mode}. The stochastic continuous-time system is governed by
\begin{align}
    \frac{\partial z}{\partial t} = (\mu + i\gamma )z - (1 + i\beta)|z|^2 z + \sqrt{\epsilon_{\mathrm{proc}}} \eta,
    \label{eq:sl}
\end{align}
where the parameters $\mu$, $\gamma$, and $\beta$ govern the limit-cycle behavior, and $\epsilon_{\mathrm{proc}}$ scales the complex-valued system noise $\eta$. The specific values for these parameters are defined in Table~\ref{tab:sl_setting}. The system is partially observed via the observation function $y_t = e^{i\theta_t}$, where $\theta_t$ denotes the phase of the complex state $z$ at time $t$.

\paragraph{Simulation, Noise Setting, and Evaluation Protocol.}
Table~\ref{tab:sl_setting} summarizes the data generation, noise settings, and evaluation protocol for the SL oscillator. Following the simulation configuration used in the ELTO supplementary setting \citep{nry2025elto}, we integrate this system numerically using the fourth-order Runge-Kutta method. To assess robustness against dynamic stochasticity, we fix the observation noise variance to $0.01$ and sweep the process noise variance $\epsilon_{\mathrm{proc}}$ from $0.01$ to $0.09$ with an increment of $0.01$. Because process noise inherently alters the Koopman spectrum, the reference spectrum used for evaluation is analytically derived from the asymptotic theory of the SL oscillator under weak process noise \citep{bag2014}. Specifically, the reference eigenvalues are computed using the fundamental frequency $\omega = 0.6$ and the decay rate parameter $\kappa = 3.0$.

\paragraph{Parameters for Our Method.}
Table~\ref{tab:sl_params} reports the architecture hyperparameters of our method used in the SL experiment. The regularization, optimization stages, and reproducibility settings are identical to those in the VDP experiment detailed in Table~\ref{tab:vdp_params}. As introduced in Appendix~\ref{subsec:aapdx-sse}, we apply the same diagonal loading for the covariance regularization to ensure numerical stability.

\end{document}